\documentclass[11pt]{article}
\usepackage[margin=1in]{geometry}
\usepackage{times}
\usepackage{microtype}
\usepackage[raggedright]{titlesec}
\usepackage[round]{natbib}
\usepackage{hyperref}
\usepackage{url}
\usepackage{amsmath,amssymb,amsthm}
\usepackage{booktabs}
\usepackage{multirow}
\usepackage{graphicx}
\usepackage{tikz}
\usepackage{enumitem}
\usetikzlibrary{positioning,fit,calc}
\usepackage{xcolor}

\newtheorem{theorem}{Theorem}
\newtheorem{proposition}{Proposition}
\newtheorem{lemma}{Lemma}

\newcommand{\smem}{SMem}

\title{Memory as a Cache:\\Exact Context Reuse and Deletion by Construction}

\author{Shengyao Wang \and Jiang Liu}
\date{}

\begin{document}
\maketitle

\begin{abstract}
The KV cache of a transformer entangles every token's representation with its entire prefix: a passage encoded once cannot be reused under a different prefix, and cannot be removed without recomputing everything after it.
Exact cache reuse in serving is therefore limited to shared prefixes, and systems that reuse more must approximate.
We present \smem{} (static memory), an architecture whose context representation is a cache by construction.
A block-local encoder maps each block of tokens to memory rows independently of all other blocks; a reader conditions generation on the union of those rows through cross-attention.
Three properties then hold for every parameter setting: memory composes exactly at fixed block indices, deleting a block is exact and is an $O(b)$ memory update for $b$-token blocks, and the memory state is independent of the edit path.
What the cache buys is measured directly.
At $4\times$ the training context, under the shared training recipe, \smem{} still retrieves planted needles beyond any window of the trained length (exact-match accuracy $0.14$--$0.28$ at distances 31 and 63 blocks) where a learned-position transformer, a RoPE transformer, a Block-Attention-style two-stream arm, and windowed or saturated readings of the transformers all score at most $0.02$; with two identical keys it returns the nearer one's value.
Because reader computation is block-local, a fully cached context is served by computing one block alone (exact up to floating-point rounding) at a near-constant $3.1$--$6.2$\,ms against the same model's cold prefill, which grows with context; batched decode holds $34$--$38\%$ fewer KV rows and runs $1.4$--$1.7\times$ faster than the transformer where decode is bandwidth-bound, and end-to-end deletion beats suffix recomputation by $8.5\times$ at 512 blocks, rising to $452\times$ at 4096 blocks; both are $32$--$256\times$ the trained length and probe the cost model rather than a served regime.
What it costs is a perplexity gap of $-4.7\%$ to $+2.8\%$ (negative favors \smem{}) against a parameter-matched transformer with the same positional scheme, at 160M--1.5B on FineWeb-Edu across two training recipes and a learning-rate search; the band's lower endpoint is set by the recipe and its upper by the learning-rate search, and the recipe's change moves the transformer twice as far as it moves \smem{} at 160M.
\smem{} also composes with RoPE: at 160M and 410M the composite leads \smem{} on every seed pair and matches or leads the matched transformer, closing $29$--$59\%$ of \smem{}'s distance to a RoPE transformer, which leads \smem{} by $7.4$--$8.5\%$.
Context representation entangled with the full prefix is a design choice; an architecture that drops that entanglement reaches comparable perplexity and gains a cache that composes, deletes, and edits exactly and, under the shared recipe, retrieves at range where every transformer variant we test does not.
\end{abstract}

\section{Introduction}
\label{sec:intro}

Prefill, the encoding of the prompt before the first output token, accounts for a substantial share of the cost of serving language models: tens of percent of GPU time in reported production workloads, and more in agentic ones dominated by long, repeated tool and document context~\citep{deepseekv3infer,llmd2025}.
The standard remedy is to cache and reuse KV states~\citep{mooncake2024}.
But a transformer's KV representation of a token is a function of its \emph{entire prefix}, so exact reuse is confined to contexts that share a prefix verbatim.
Hit rates are high for chat-style prefix repetition and low for retrieval, multi-agent, and re-ordered document workloads: one production study finds only $8\%$ of requests and $18\%$ of prefill tokens covered by prefix reuse~\citep{cachecraft2025}; the reusable-but-non-prefix mass (retrieved passages, tool schemas, shared documents) is structurally out of reach.
A growing line of work therefore reuses KV states beyond exact prefix match---recomputing, blending, or compressing a fraction of them~\citep{promptcache2024,cacheblend2024,blockattention2024,recache2026}.
The limitation is intrinsic: the cached quantity was never designed to compose, so every reuse carries an approximation error that must be measured and re-measured per deployment; independent evaluations report quality drops of $7$--$18$ points against full prefill~\citep{cacheblendaudit2026}, motivating verification layers~\citep{vericache2026}.

Prefix entanglement blocks removal for the same reason: evicting one passage (a retracted document, a stale tool result, data a user asked to remove) invalidates the KV states of everything after it, so the system must recompute the suffix or leave the content in place.

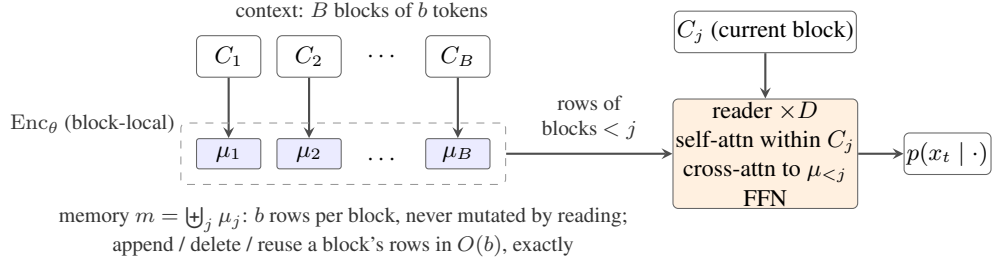
\begin{figure}[t]
\centering
\begin{tikzpicture}[
  font=\footnotesize,
  blk/.style={draw=black!60, rounded corners=2pt, minimum width=8.5mm, minimum height=5.5mm, inner sep=1pt},
  mem/.style={draw=black!60, fill=blue!8, rounded corners=1pt, minimum width=8.5mm, minimum height=4mm, inner sep=1pt},
  op/.style={-stealth, black!70, thick},
  lbl/.style={black!80, font=\scriptsize}
]
\node[blk] (c1) {$C_1$};
\node[blk, right=2mm of c1] (c2) {$C_2$};
\node[right=2mm of c2] (cd) {$\cdots$};
\node[blk, right=2mm of cd] (cb) {$C_B$};
\node[lbl, above=1mm of c2.north, xshift=7mm] {context: $B$ blocks of $b$ tokens};
\node[mem, below=8mm of c1] (m1) {$\mu_1$};
\node[mem, below=8mm of c2] (m2) {$\mu_2$};
\node[below=9.5mm of cd] (md) {$\cdots$};
\node[mem, below=8mm of cb] (mb) {$\mu_B$};
\draw[op] (c1) -- (m1);
\draw[op] (c2) -- (m2);
\draw[op] (cb) -- (mb);
\node[lbl, left=1.5mm of m1.west, yshift=4mm, anchor=east] {$\mathrm{Enc}_\theta$ (block-local)};
\node[draw=black!40, dashed, rounded corners=2pt, fit=(m1)(mb), inner sep=2mm] (memall) {};
\node[lbl, below=1.5mm of memall.south, align=center, anchor=north] {memory $m=\biguplus_j\mu_j$: $b$ rows per block, never mutated by reading;\\append / delete / reuse a block's rows in $O(b)$, exactly};
\node[blk, fill=orange!12, right=22mm of memall.east, minimum width=21mm, minimum height=12mm, align=center, anchor=west] (reader) {reader $\times D$\\self-attn within $C_j$\\cross-attn to $\mu_{<j}$\\FFN};
\node[blk, above=6mm of reader.north, minimum width=14mm, anchor=south] (cj) {$C_j$ (current block)};
\draw[op] (cj) -- (reader);
\draw[op] (memall.east) -- node[lbl, above=0.5mm, align=center] {rows of\\blocks $<j$} (reader.west);
\node[blk, right=6mm of reader.east, minimum width=11mm, anchor=west] (out) {$p(x_t\mid\cdot)$};
\draw[op] (reader) -- (out);
\end{tikzpicture}
\caption{\smem{}. Each block is encoded independently (attention masked to the block); the reader attends within the current block and cross-attends into the memory rows of earlier blocks. Composition, deletion, and path independence hold for every $\theta$.}
\label{fig:arch}
\end{figure}

This paper asks the architectural question instead: \emph{what if the model's context representation were a cache by design?}
We propose \smem{} (static memory: a block's rows never change once encoded).
Context is split into blocks; an encoder with block-local attention maps each block to memory rows \emph{independently of every other block}; generation conditions on the union of memory rows through cross-attention (Figure~\ref{fig:arch}).
Because a block's memory never depends on its neighbors, three properties hold for all parameters $\theta$---they are consequences of the connectivity pattern, not of training:
\begin{itemize}
\item \textbf{Exact composition.} The memory of any block set is the union of their individually encoded memories: a block encoded once is reusable under any surrounding content at its encoded index, with zero recompute and no approximation error; moving it to a \emph{different} index costs one $O(b)$ re-encode of that block alone.
\item \textbf{Exact deletion.} Removing a block's rows from memory is an $O(b)$ update for block size $b$, independent of memory size, and yields \emph{exactly} the state in which the block was never encoded (later blocks keep their indices; no re-indexing is implied).
\item \textbf{Path independence.} Any sequence of insertions and deletions arriving at the same block set yields the same memory and the same outputs; sequential edits do not accumulate error.
\end{itemize}
A transformer obtains them only by re-encoding the context, at $O(n)$ rather than $O(b)$; \smem{} moves the cross-block conditioning out of encoding and into the reader, over a memory that reading never modifies, so the composed memory is the model's own semantics and there is no reconstruction error to bound (cf.~\citealp{linearkv2026}).

The nearest designs are encoder--decoder memories.
RETRO~\citep{retro2022}, CEPE~\citep{cepe2024}, and YOCO~\citep{yoco2024} read chunk-independent encodings from decoders that self-attend over the full prefix, so their chunks compose but their reader state does not (a CEPE-style reader on \smem{}'s memory trails it by $9.7\%$ at matched parameters and depth split, Appendix~\ref{app:recipe}); COMB~\citep{comb2026} retrofits a chunk-local encoder and cross-attention onto a frozen decoder explicitly for position-independent caching and reports TTFT reductions of $51$--$94\%$ at comparable accuracy, and as with CEPE its chunks compose but its reader state does not, with no stated guarantee of exactness, deletion, or path independence; Block-Attention~\citep{blockattention2024} retrofits block-local KV into a pretrained transformer, and our from-scratch arm of that design reaches perplexity parity with zero retrieval at range and neither deletion nor path independence (\S\ref{sec:niah}).
Compressive and streaming memories bound the state by summarizing or evicting context~\citep{infiniattn2024,autocomp2023,streamingllm2024}; \smem{} keeps $b$ uncompressed rows per block so that composition and deletion stay exact, at memory linear in context.
Larimar~\citep{larimar2024}, Memory Mosaics~\citep{mosaics2024}, removable kNN memory~\citep{munkey2026}, reversible external-memory unlearning~\citep{ke2026}, and audit-conditional deletion for retrofitted memories~\citep{ramesh2026} pursue adjacent goals without parameter-independent guarantees, and surveys of memory architectures omit deletability as a classification axis~\citep{memsurvey2026a,memsurvey2026b}.
The unlearning literature measures how hard removal is when it is not architectural: relearning attacks recover unlearned content~\citep{relearn2025}, deleted documents resurface under retrieval~\citep{ragdel2026}, localized parameter edits damage unrelated knowledge~\citep{smft2025}, and unlearning now carries retraining-equivalence obligations on weights~\citep{overused2026}; what \smem{} provides is removability by construction for context-borne knowledge, with the weight/memory split measured in \S\ref{sec:delete}.

\textbf{Contributions.}
\begin{itemize}[leftmargin=1.2em, itemsep=0pt, topsep=1pt]
\item An architecture whose memory composes, deletes, and edits exactly for every parameter setting, verified numerically on random parameters and on trained checkpoints (\S\ref{sec:arch}).
\item Retrieval at $4\times$ the trained length under the shared recipe (exact match $0.14$--$0.28$ at distances 31 and 63 blocks) where every transformer variant we test scores at most $0.02$ (\S\ref{sec:cache}).
\item An exactness-verified block-skip path that serves cached contexts at near-constant cost, bandwidth-bound decode $1.4$--$1.7\times$ faster on $34$--$38\%$ fewer KV rows, and delete-then-generate $8.5$--$452\times$ faster than suffix recomputation (\S\ref{sec:cache}).
\item Matched-parameter pretraining at 160M--1.5B: perplexity comparable to the matched transformer's, a gap of $-4.7\%$ to $+2.8\%$ across two recipes and a learning-rate search, and an \smem{}+RoPE composite that beats \smem{} at 160M and 410M and matches or beats the matched transformer (\S\ref{sec:pretraining}).
\item A mechanism account: a $+0.16$-nat encoding tax concentrated $8.8\times$ on tokens whose evidence sits in an earlier block, absorbed by reader layers, and a synthetic chain task whose reader-round requirement is bounded above by an explicit construction and met exactly by trained models (\S\ref{sec:mechanism}).
\end{itemize}

\section{Architecture and guarantees}
\label{sec:arch}

\paragraph{Model.}
The input is partitioned into $B$ blocks $C_1,\dots,C_B$ of $b$ tokens ($n=Bb$).
A single encoder of $E$ layers, applied to every block with attention masked to within-block (bidirectional inside the block), produces memory rows
$\mu_j=\mathrm{Enc}_\theta(C_j)\in\mathbb{R}^{b\times d}$, regarded as the multiset of its $b$ rows; positions are assigned from the block's global index at encode time, so a cached $\mu_j$ carries its own addressing (the \smem{}+RoPE composite of \S\ref{sec:pretraining} instead carries addressing in the reader).
The memory is the multiset $m=\biguplus_{j} \mu_j$, stored canonically ordered by block index; it is never mutated during reading.
The reader is a stack of $D$ decoder layers.
Each applies causal self-attention masked to the position's \emph{own block} (never across a block boundary), then cross-attention into the memory rows of \emph{strictly earlier} blocks under a block-causal mask, with a learned null row so a query may abstain, then an FFN; the final state gives the next-token distribution.
The memory holds $b$ uncompressed rows per $b$-token block, and the encoder is residual, so each token's embedding enters its row additively (before row normalization): block-local encoding constrains \emph{access} to information, not its presence.
Beyond the trained block count, indices are assigned by a saturating \emph{slot scheme}: the most recent $B$ blocks receive indices $1,\dots,B$ as at training length and everything older is folded into slot $1$ (an append then re-indexes the $B{-}1$ most recent blocks, so beyond $B$ blocks exact reuse and retrieval at range do not coincide; the rotary composite of \S\ref{sec:pretraining} has both; Appendix~\ref{app:arch}).

\paragraph{Structural guarantees ($\forall\theta$).}
The following hold for every parameter setting, verified numerically to machine precision on random parameters (gates G1--G8 on an fp64 reference implementation, Appendix~\ref{app:gates}) and on the trained checkpoints through the block-skip and deletion gates (Appendices~\ref{app:serving} and~\ref{app:niah}).

\begin{proposition}[Exact composition]
\label{prop:comp}
At fixed block indices, $\mathrm{Enc}_\theta$ applied to a concatenation equals the multiset union of per-block encodings: writing $\mu_{1:2}$ for the joint encoding, $\mu_{1:2}(C_1\Vert C_2)=\mu_1(C_1)\uplus\mu_2(C_2)$ exactly, each block encoded at its own index. \emph{(The attention mask makes each block's computation a function of that block and its index alone.)}
\end{proposition}

\begin{theorem}[Exact editing]
\label{thm:edit}
Memories under $\uplus$ form a cancellative commutative monoid, so deletion of a present block is well defined; deleting $\mu_j$ yields exactly the memory in which $C_j$ was never encoded at index $j$ (remaining blocks keep their indices), and the resulting memory depends only on the current block multiset, not on the edit path. Every read inherits this: cross-attention is permutation-invariant over memory rows and block indices are frozen at encode time, so reads are a function of the multiset alone.
\end{theorem}

The edit itself touches only the $b$ rows of the affected block; the storage layout sets where the bookkeeping cost lands (a contiguous layout shifts subsequent rows at delete time; an index-mapped layout keeps deletion flat but pays a read-time gather that grows with memory---both measured, Appendix~\ref{app:serving}).
Path independence makes the degradation curve under sequential editing exactly flat, with no error accumulation to bound; memories and outputs are bit-identical over 100 random edit orders (Appendix~\ref{app:gates}).

\begin{lemma}[Cross-attention sensitivity is independent of memory size at fixed diameters]
\label{lem:jac}
For one query input $q$ (query vector $W_Q^{\top} q$) with attention weights $a$ over memory rows, the Jacobian of the cross-attention output $c=\sum_i a_i v_i$ is $\partial c/\partial q=\frac{1}{\sqrt d}\,\mathrm{Cov}_a(v,k)\,W_Q^{\top}$ with $\mathrm{Cov}_a(v,k)=\sum_i a_i (v_i-\bar v)(k_i-\bar k)^{\top}$, and
$\Vert\partial c/\partial q\Vert_2\le \mathrm{diam}(K)\,\mathrm{diam}(V)\,\Vert W_Q\Vert_2/(4\sqrt d)$,
independent of the number of rows at fixed key/value diameters.
\end{lemma}

The bound's memory dependence enters only through the diameters, which row normalization bounds analytically (each row is $\gamma\odot\hat x+\beta$ with $\Vert\hat x\Vert_2=\sqrt d$, so the diameters depend on the affine parameters and the projections, not on the row count; checked across a $16\to4096$-row sweep, Appendix~\ref{app:gates}); it bounds sensitivity, not retrieval fidelity, which \S\ref{sec:niah} measures.

\paragraph{Expressivity and the cost of reading.}
Two results frame what the reader must supply: the first concerns the reader's layer form, the second a synthetic testbed whose reader is a weight-tied recurrence with $R$ rounds and $w$ query slots (the LM reader has $D$ untied layers and per-position states in those roles).
First, the reader class contains the transformer as a limiting case: with self-attention, cross-attention, and FFN in sequential pre-LN residual branches as in the LM reader, at $w=n$, $R=L$ ($L$ the transformer's layer count), untied per-round weights, an unrestricted attention window, and zeroed cross-attention, one round is a standard transformer layer (Appendix~\ref{app:theory}).
Second, for a task family requiring a $k$-step retrieval chain, a round-elimination argument predicts $R=\Omega(k)$ for a narrow reader, and the upper direction is an explicit construction: weights solving the task with $R=k$ at \emph{any} width, exactly, verified on 105 width/size/depth cells (Appendix~\ref{app:theory})---giving, under the capacity assumption stated there, a per-width floor of $R=k$, flat in $w$, below a width-substitution threshold.
\S\ref{sec:mechanism} tests both against trained models.

\section{The cache in operation}
\label{sec:cache}

Unless noted, measurements use checkpoints from the shared-recipe runs (Regime 1); \S\ref{sec:pretraining} contrasts that recipe with a tuned one (Regime 2), on which the retrieval separation replicates within the trained window and thins beyond it; serving benchmarks run on a single H200 (decode on an L40S), bf16, batch and context stated per cell (Appendix~\ref{app:serving}).

\subsection{Reuse: an exact fast path for cached contexts}
\label{sec:reuse}

\begin{figure}[t]
\centering
\includegraphics[width=\linewidth]{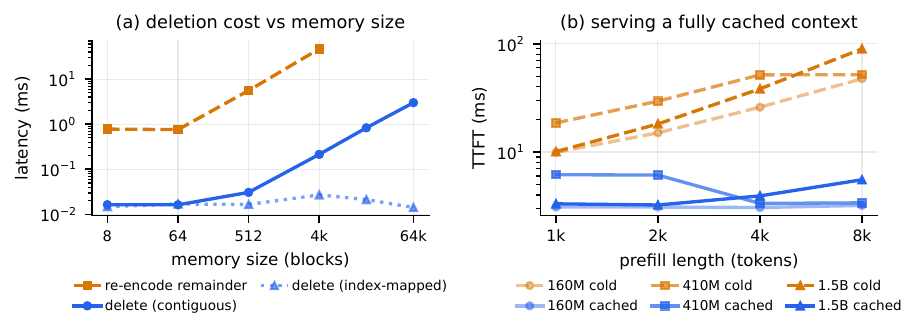}
\caption{(a) Deletion latency vs memory size: re-encoding the remainder (pilot-scale model), contiguous deletion, and index-mapped deletion (flat; its read-time gather cost is in Appendix~\ref{app:serving}). (b) Serving a fully cached context by block-skip vs cold prefill (410M's 8k prefill point is an unresolved anomaly).}
\label{fig:cache}
\end{figure}

Composition (Proposition~\ref{prop:comp}) makes reuse exact, and the reader's block-locality makes it cheap to \emph{serve}: a position's reader state depends only on its own block's tokens and the (cached) memory, so a fully cached context is served by running the reader on the final block alone.
This block-skip path is not an approximation---its final-block logits equal the full reader pass up to floating-point rounding (max abs $4.7\times10^{-5}$, argmax agreement $100\%$, all scales and lengths tested).
Its cost is near-constant at $3.1$--$6.2$\,ms per request across context lengths (the pass is $O(bn)$ in cross-attention, linear in context, against the full pass's $O(n^2)$; the flatness is the launch-bound regime), while the model's own cold prefill grows with context (Figure~\ref{fig:cache}b): a $67$--$69\%$ time-to-first-token (TTFT) reduction at training length and, mechanically, more at longer contexts ($88$--$94\%$ at $4$--$8\times$).
The encoder is $15$--$22\%$ of prefill FLOPs; under partial reuse the reader runs from the first uncached block.

Decode inherits the KV economics.
For a served context of $n$ tokens, the reader holds cross-attention K/V over the memory plus only $b$ rows of block-local self-attention KV per layer, across $D<L$ layers: $D(n{+}b)$ vs $L\cdot n$ rows, i.e.\ $34$--$38\%$ fewer KV rows, hence bytes (head dimension and dtype match), than the learned-position transformer at our depth splits (the driver is $D<L$, a depth split a transformer could take at a quality cost, Appendix~\ref{app:recipe}; grouped-query attention composes with it; Appendix~\ref{app:acct}).
Where decode is bandwidth-bound (batch 16--64, both arms at ${\ge}590$ GB/s on an L40S), the fewer bytes convert to $1.4$--$1.7\times$ higher throughput at $4$--$18\%$ lower peak memory, in line with the row ratio; at batch 1 both arms are launch-bound and within $1.21\times$ (exactness-gated, contiguous layout; Appendix~\ref{app:serving}).
Recomputing K/V from the memory rows instead of storing it cuts peak memory further at $0.1$--$0.2\times$ the throughput (Appendix~\ref{app:serving}).

Prefix caching saves prefill only for prefix-identical requests ($8\%$ in the production study above); \smem{}'s saving applies to any repeated block under any surrounding content, with no reconstruction error.

\subsection{Retrieval that survives length under the shared recipe, unlike every transformer variant tested}
\label{sec:niah}

Extending the context appends memory rows under the slot scheme rather than stretching a positional pattern.
We test with a needle design under paired controls: plant an 8-token key with a 1-token value in FineWeb-Edu validation text at a controlled block distance from a query at the end.
We measure exact-match accuracy (argmax at the answer position) and conditional gain (matched-key minus mismatched-key log-probability of the value, canceling presence bias); $n{=}384$ or $1024$ per cell, by harness; protocol in Appendix~\ref{app:niah}.

\begin{figure}[t]
\centering
\includegraphics[width=0.7\linewidth]{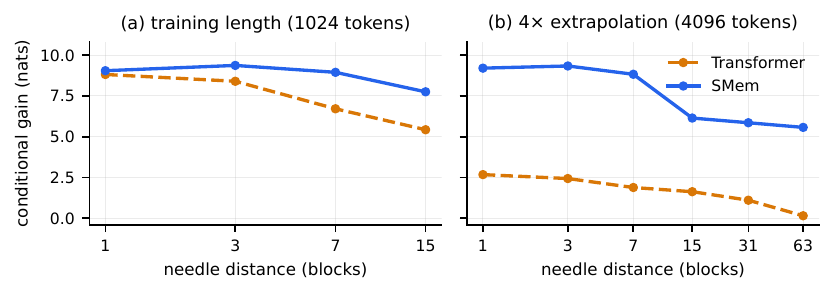}
\caption{Needle retrieval at 1.5B: conditional gain by needle distance. At $4\times$ the training length the transformer's signal collapses; \smem{}'s survives to the farthest distance tested (grid and controls in Appendix~\ref{app:niah}).}
\label{fig:niah}
\end{figure}

At $4\times$ the training length, \smem{} retrieves with exact-match accuracy $0.14$--$0.28$ at distances 31 and 63 blocks across scales, beyond any window of the trained length ($0.13$--$0.26$ at distance 15, which a window reaches; primary harness; Figure~\ref{fig:niah} shows the conditional-gain view), while the learned-position transformer reads $0.00$--$0.02$.
In distribution, at the training length and distance 15 (160M, primary harness): \smem{} $0.52$, learned-position transformer $0.15$, RoPE transformer $0.00$; the RoPE zero reflects our 3.2B-token budget (four runs; distance-1 control $0.57$--$0.64$): a pretrained Pythia-410M reads $0.64$ in distribution on the same probe and $0.00$ at $2\times$ its context under direct extension (Appendix~\ref{app:niah}).
Block-locality alone does not explain it: the Block-Attention-style arm reads $0.00$ everywhere at $4\times$ (conditional gain $\le0.35$ nats) while matching the learned-position baseline in distribution.
Extensions of the transformer arms' own (RoPE offsets clamped; position interpolation, NTK-aware scaling, and YaRN at inference time; learned positions clipped) recover at most part of the near range and nothing beyond 15 blocks, and a sliding window over the last $T$ tokens, with or without attention-sink tokens, keeps in-distribution accuracy up to 15 blocks and reads $0.00$ beyond (Table~\ref{tab:niah}; Appendix~\ref{app:niah}).

\begin{table}[t]
\centering
\caption{Exact-match retrieval at $4\times$ the trained length by needle distance in blocks (160M unless noted, replication harness, $n{=}384$ per cell; ranges span seeds and regimes as listed; sat.\ = block offsets saturated at the trained range). Cross-scale primary-harness numbers are in the text; full grid in Appendix~\ref{app:niah}.}
\label{tab:niah}
\vspace{2pt}
\footnotesize
\setlength{\tabcolsep}{4pt}
\begin{tabular}{llcccccc}
\toprule
arm & runs & 1 & 3 & 7 & 15 & 31 & 63 \\
\midrule
\smem{} & R1 s0 & .71 & .57 & .42 & .18 & .13 & .12 \\
\smem{} & R2, 3 seeds & .46--.74 & .24--.65 & .13--.43 & .01--.13 & .01--.11 & .01--.11 \\
\smem{}+RoPE, sat. & R1+R2, 6 runs & .65--.73 & .47--.60 & .24--.42 & .07--.11 & .03--.12 & .05--.10 \\
\smem{} & 410M R2, 2 seeds & .55--.60 & .38--.42 & .13--.20 & .01--.05 & .00--.04 & .00--.03 \\
\smem{}+RoPE, sat. & 410M R2 s0 & .75 & .77 & .59 & .15 & .14 & .14 \\
\midrule
transformer (learned pos.) & R1 s0, R2 3 seeds & .00--.02 & .00--.01 & .00 & .00--.01 & .00--.01 & .00 \\
\quad + window of last $T$ & 4 runs & .46--.69 & .33--.52 & .26--.45 & .15--.28 & .00 & .00 \\
RoPE transformer & R1 3 seeds, R2 s0 & .00 & .00 & .00 & .00 & .00 & .00 \\
\quad + YaRN (best extension) & 4 runs & .04--.08 & .02--.04 & .05--.09 & .04--.14 & .00--.02 & .00 \\
\quad + sink + window & 4 runs & .56--.66 & .33--.54 & .06--.26 & .00 & .00 & .00 \\
Block-Attention arm & R1, 3 seeds & .00 & .00 & .00 & .00 & .00 & .00 \\
\bottomrule
\end{tabular}
\end{table}

Two alternative explanations fail.
The recipe: on tuned-recipe checkpoints it replicates ($0.13$--$0.74$ at distances $\le7$ vs ${\approx}0.00$ for the transformer), with accuracy beyond the window ($d{\ge}31$) of $0.00$--$0.11$ across seeds and scales, near the floor at 410M, against $0.12$--$0.13$ for the shared-recipe 160M seed in that harness, while the margin in conditional gain stays at $1.2$--$3.2$ nats against $\le0.5$: the tuned recipe buys perplexity and thins the far-range signal.
The positional scheme: an \smem{} variant whose positions come from rotary phases rather than a table (\S\ref{sec:pretraining}) also retrieves at range once its block offsets are capped at the trained range, as the slot scheme caps indices ($0.65$--$0.75$ at distance 1, $0.03$--$0.14$ at $\ge15$ blocks over six 160M runs and one 410M run; at 410M under the tuned recipe it reads $0.14$ at distances 31 and 63 against \smem{}'s $0.00$--$0.04$), at a near-range cost against a windowed reading of the same model ($0.09$--$0.11$ vs $0.33$--$0.34$ at distance 15 on the same two seeds; Appendix~\ref{app:niah}).
Retrieval is also structured rather than bag-of-keys: with two identical keys at different distances, \smem{} returns the nearer key's value ($0.65$ vs $0.01$ at 160M; $0.50/0.11$ at 1.5B; Appendix~\ref{app:niah}).

\subsection{Deletion as cache management}
\label{sec:delete}

Theorem~\ref{thm:edit} and gate G7 establish that the post-deletion memory \emph{is} the never-encoded memory; the empirical question is how much of a behavior the memory carried.
In a controlled setting (small-scale models with bidirectional reading, so own-block memory exists, trained on in-context copying of rare tokens; Appendix~\ref{app:suppress}), deleting every block containing the evidence for a target drops the target's accuracy from $0.58$--$0.61$ to $0.361$--$0.363$, at the same models' own-block-only floor ($0.365$), while a locally-predictable control stratum changes by at most $0.001$.
On the pretrained models, deleting the needle's block returns retrieval to the never-planted rate in all 40 cells (exact match $\le0.003$ from $0.03$--$0.82$; deleting a neighboring block changes it by less than $0.04$; Appendix~\ref{app:niah}).
Because the memory update is exact, there is no residual \emph{in memory}; of the $0.60$ ability, $0.23$ is the cross-block share deletion removes, $0.18$ own-block memory, $0.19$ the weights and own-block tokens (Appendix~\ref{app:suppress}).

End to end, deletion composes with the block-skip path: delete the rows, then generate.
The flat deletion curve is the index-mapped layout and the decode gain of \S\ref{sec:reuse} the contiguous one; no layout in our implementation gives both (Appendix~\ref{app:serving}).
The comparator recomputes the transformer's suffix KV, since masking KV in place does not yield the never-encoded state.
The delete-then-generate cycle is $8.5\times$ faster in the most modest cell and $452\times$ in the most favorable (160M, 4096 blocks, deletion nearest the start); at 1.5B with 4096-block contexts it is $125$--$126$\,ms vs $13.9$--$29.9$\,s (per-cell ratios $111$--$237\times$; Appendix~\ref{app:serving}).

\section{What the move costs: matched-parameter pretraining and a recipe--\allowbreak architecture interaction}
\label{sec:pretraining}

\begin{table}[t]
\centering
\caption{Validation perplexity on FineWeb-Edu (means over seed-paired streams; per-seed statistics in Appendices~\ref{app:perseed}/\ref{app:recipe}). Left: the matched pair (learned positions; $\Delta$ = \smem{} vs transformer) at every scale and regime run. Right: the positional-scheme arms (RoPE transformer; \smem{}+RoPE composite), to be read against the matching rows on the left; 160M rows are three-seed means, the 410M row a single seed.}
\label{tab:main}
\vspace{2pt}
\setlength{\tabcolsep}{4pt}
\begin{tabular}{llccc}
\toprule
scale & regime & transf. & \smem{} & $\Delta$ \\
\midrule
160M & 1 (no warmup) & 29.72 & 28.32 & $-4.7\%$ \\
160M & 2 (warmup) & 26.51 & 26.96 & $+1.7\%$ \\
410M & 1 (no warmup) & 21.26 & 20.32 & $-4.4\%$ \\
410M & 2 (warmup) & 19.10 & 19.50 & $+2.1\%$ \\
1.5B & 1 (no warmup) & 14.46 & 14.26 & $-1.4\%$ \\
\bottomrule
\end{tabular}\hspace{6mm}%
\begin{tabular}{lcc}
\toprule
scale, regime & RoPE & +RoPE \\
\midrule
160M, 1 & 26.11 & 27.01 \\
160M, 2 & 25.10 & 26.30 \\
410M, 2 & 18.25 & 19.26 \\
\bottomrule
\end{tabular}
\end{table}

We ran the matched comparison under two training recipes: the one our pilots selected (Regime 1) and one re-tuned on the transformer arm after a registered control on the recipe axis exceeded its decision threshold (Regime 2).

\paragraph{Setup.}
\smem{} and a standard pre-LN transformer are pretrained from scratch on FineWeb-Edu~\citep{fineweb2024} (GPT-2 tokenizer, $T{=}1024$, $b{=}64$, $B{=}16$) at ${\approx}20{:}1$ token budgets~\citep{chinchilla2022}: 160M/3.2B tokens (3 seeds per arm), 410M/8.2B (2 seeds), 1.4B/30B (1 seed; we write ``1.5B'' for the $1.42$B-parameter scale).
The seed fixes the data order and the evaluation stream for both arms, so per-seed \emph{paired differences} are the estimand (streams differ across seeds; displayed means average them).
Parameters match within $1.5\%/0.01\%/0.00\%$ at the three scales (\smem{} higher) and analytic FLOPs/token within $4.4\%/4.6\%/2.5\%$ (\smem{} lower, from block-local self-attention and the shallower reader), while our implementation trains $15$--$24\%$ slower per GPU in wall-clock (Appendix~\ref{app:acct}).
Causality is verified by exact-zero leak tests at every scale (Appendix~\ref{app:leak}).
A per-run control shows the memory is read: replacing each sequence's memory with another sequence's from the same batch (content wrong, row statistics intact) costs $0.83$--$1.07$ nats in Regime 1 and $0.85$--$0.97$ in Regime 2; we call this the \emph{roll gap} (Appendix~\ref{app:roll}).
Throughout, the \emph{retrievable-rare} stratum is the set of target tokens of train-corpus frequency rank $\ge128$ whose exact token occurs in a strictly earlier block: the tokens for which cross-block access should matter.

\paragraph{Regime 1, and a registered prediction.}
A width ladder of pilots ($d\in\{128,256,384\}$) showed the \smem{} deficit shrinking monotonically with width ($6.3\%\to4.4\%\to2.9\%$).
Before any run here started we registered the extrapolation (a deficit of at most ${\sim}2\%$ at 410M), with the 410M run as the held-out test and a kill criterion on the 160M run that did not bind.
The outcome inverted the prediction's sign: \smem{} reached \emph{lower} perplexity at all three scales (Table~\ref{tab:main}; 160M $-4.7\%$, paired $t(2){=}-25.2$, $p{=}0.002$; 410M $-4.4\%$, sign-consistent, paired $p{=}0.11$ at $n{=}2$; 1.5B $-1.4\%$ on the single pair).

\paragraph{Regime 2: a registered recipe control.}
The shared recipe has no learning-rate warmup, a choice fixed during \smem{} pilots; we therefore registered a control with a decision rule written before its trigger runs: give the 410M transformer standard warmup (2000 of 15641 steps), and separately a no-warmup higher-LR variant ($4.5{\times}10^{-4}$), re-basing the comparison if either gained more than $1\%$.
Both did ($-9.4\%$ and $-6.1\%$; the LR-only variant shows the effect is not specific to warmup).
Both cleared the re-base bar, so we re-trained the comparison under the tuned recipe at 160M (3 seeds per arm, ``3v3'') and 410M (2v2); schedules and corpus are identical across arms within each regime, and the gap is already present in training loss (Appendix~\ref{app:recipe}).
Under Regime 2 the transformer leads: $+1.7\%$ at 160M (paired $t(2){=}36.3$, $p{<}0.001$, CI $[+1.5\%,+1.9\%]$) and $+2.1\%$ at 410M (per-seed $+0.418/+0.395$ ppl; a paired test on two seeds, $p{=}0.018$, is descriptive only).
Roll gaps under warmup stay at $0.85$--$0.97$ nats, so Regime 2 is a valid operating point for both arms.

\begin{figure}[!b]
\centering
\includegraphics[width=0.5\linewidth]{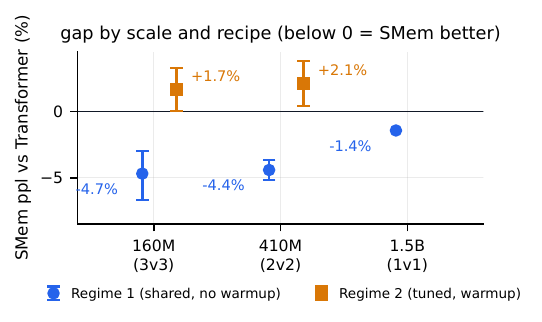}
\caption{Perplexity gap (\smem{} vs learned-position transformer) by scale and regime; markers are paired means, bars span cross-seed combinations (CIs in Appendix~\ref{app:perseed}; 1.5B is a single Regime-1 pair).}
\label{fig:scaling}
\end{figure}

\paragraph{Scope of the perplexity claim.}
Across regimes and scales the gap stays inside a $-4.7\%$ to $+2.1\%$ band ($+2.8\%$ with Appendix~\ref{app:recipe}'s learning-rate search; Figure~\ref{fig:scaling}): the architectures reach comparable perplexity, and the recipe axis moves the transformer further than the architecture axis separates the arms ($10.2\%$ recipe effect on the transformer at 410M against a $2.1$--$4.4\%$ architecture contrast; \smem{}'s own recipe effect is $4.0$--$4.8\%$).
The interaction is asymmetric ($10.8\%$ vs $4.8\%$ at 160M; $3.9\%$ for the RoPE transformer, so the sensitivity tracks the learned-position table rather than block-locality), and since each recipe was fixed on one arm we report the band across both rather than an endpoint; a symmetric learning-rate search at 160M moves both arms by $10$--$11\%$ and leaves the contrast at $+2.4$ to $+2.8\%$ (Appendix~\ref{app:recipe}).
Stratified, \smem{} leads the retrievable-rare stratum in Regime 1 at all scales and holds parity or better on it under Regime 2, so the tuned-recipe deficit lies outside the retrieval stratum (Appendix~\ref{app:perseed}).
Two controls close alternative explanations for Regime 1: a depth-matched transformer (\smem{}'s layer count, widened FFN, exact parameter match) is \emph{worse} than the standard baseline (gap closure $-27\%$), and the Block-Attention-style arm lands at baseline parity; under Regime 2 a CEPE-style reader with full-prefix self-attention over the same memory, at the same parameters and depth split, trails \smem{} by $9.7\%$ and matches a memory-free 7-layer transformer (Appendix~\ref{app:recipe}).
Zero-shot evaluations (Regime-1 checkpoints) are at parity on four of five tasks; LAMBADA favors \smem{} at every scale ($+0.027$--$0.029$, about three item-sampling standard errors; Appendix~\ref{app:zeroshot}).

\paragraph{\smem{} composes with RoPE.}
Nothing forces memory rows to carry absolute position: rotary phases can instead be applied inside blocks and to the block \emph{offset} in cross-attention.
The resulting \smem{}+RoPE composite ($27.01$/$26.30$ at 160M, three seeds per recipe) leads both the matched transformer and \smem{} on every seed pair in both regimes, keeps retrieval at range (\S\ref{sec:niah}), and closes $59\%$/$35\%$ of \smem{}'s distance ($+8.5\%$/$+7.4\%$) to a RoPE transformer trained under the same recipes ($26.11/25.10$ at 160M, the strongest arm on validation perplexity).
At 410M under Regime 2 (single seed) the same composite reads $19.26$: level with the learned-position transformer ($19.25$), $2.1\%$ below \smem{} ($19.67$), and $5.5\%$ above a RoPE transformer at $18.25$, which under Regime 1 reads $18.89$; the RoPE arm's margins ($5.5\%$/$12.4\%$ over the learned-position transformer, $7.8\%$/$8.3\%$ over \smem{} under Regimes 2/1) and the composite's $29\%$ closure carry the 160M pattern to 410M (Appendix~\ref{app:recipe}).

\section{Mechanism: the encoding tax and the price of reading}
\label{sec:mechanism}

\paragraph{The encoding tax is real and localized.}
A block-local encoder cannot compute features conditionally on other blocks: it must encode unconditionally whatever might later be needed.
We measure this with from-scratch matched pairs (parameter-identical ${\approx}83.6$M decoder-only models differing only in their attention mask, block-local-causal vs full-causal; a decoder-only proxy cannot use \smem{}'s within-block bidirectional mask without exposing targets, so the pair isolates the cross-block part of the tax; 1B tokens, three seed pairs): the block-local model pays $+0.156\pm0.034$ nats overall (mean $\pm$ SD over pairs), concentrated on the stratum that argument implicates: $+1.34\pm0.06$ on the retrievable-rare stratum, a mean per-pair concentration of $8.8\times$ (range $7.2$--$10.0$), while the non-retrievable-rare stratum moves slightly the other way in every pair ($-0.14\pm0.04$; Appendix~\ref{app:tax}).
A mask-swap probe on a pretrained Pythia-410M~\citep{pythia2023}, which never adapted to the mask, gives an upper bound with the same concentration ($+3.04$ vs $+0.57$ nats).
At pilot scale a readerless block-local model pays a $15$--$21$-point deficit on cross-block-dependent predictions that does not shrink with encoder width, disappears once both arms read through four rounds, and is not absorbed by equal-FLOPs deepening of the encoder: read rounds do the work (Appendix~\ref{app:tax}).

\paragraph{Reading has a measurable resource relation, closed from above and consistent from below.}
On the $k$-step retrieval-chain testbed (permutation domain $n'{=}32$), the construction of \S\ref{sec:arch} gives $R=k$ at any width from above; from below, every formed cell (a cell where training reached the task) at $w\le4$ succeeds at $R=k$ and the per-seed minimal $R\cdot w$ equals $k$ exactly (attained at $w{=}1$, $R{=}k$); the grid searched $R\ge k$ at $w\le4$, and sub-$k$ rounds were run only at $w\ge8$, where width substitution appears (Appendix~\ref{app:synth}).
Aggregate scaling is consistent: minimal $R\cdot w$ grows roughly linearly in $k$ (OLS exponent $1.27$, 95\% CI $[0.98,1.56]$; rank-based estimate $1.00$; robustness and censoring sensitivities in Appendix~\ref{app:synth}).
Width substitution appears only at $w\ge8$ ($=n'/4$), where the capacity assumption behind the floor no longer holds and whole-table copying becomes available (Appendix~\ref{app:theory}).
The measurement relates to communication-complexity bounds for attention~\citep{sanford2024} and iteration--capacity exchange~\citep{iterlaw2026}; unlike test-time-depth models, which report mixed returns on added iterations~\citep{cart2026}, ours is fixed-$R$ absorption of the encoding tax during training.

\section{Limitations}
\label{sec:limits}

\smem{} is a pretraining-time choice, not a retrofit.
The perplexity gap to the matched transformer spans $-4.7\%$ to $+2.8\%$; the RoPE transformer leads all arms, and the composite that closes $29$--$59\%$ of the distance to it has a single seed at 410M and none at 1.5B; evidence at 1.5B is a single seed pair under Regime 1, and under the tuned recipe the far-range exact-match signal sits near the floor (\S\ref{sec:niah}), so a deployment that tunes for perplexity should not assume it.
Pretraining uses $T{=}1024$ with 16 blocks; beyond that, appends re-encode $B{-}1$ blocks, and the decode gain and the flat deletion curve are measured in different memory layouts (Appendix~\ref{app:serving}).
The retrieval margin rests on needle, duplicate-key, and natural-text probes; open-book QA is harness-sensitive (Appendix~\ref{app:niah}).
Deletion is exact for context held in memory and says nothing about the weights, and our training implementation runs $15$--$24\%$ slower per GPU than the baseline (Appendix~\ref{app:acct}).

\section{Conclusion}

An architecture that encodes blocks independently and pays for cross-block conditioning at read time reaches perplexity comparable to a parameter-matched transformer's at every scale-and-recipe pair we ran, and gains a cache that composes and deletes exactly, serves cached contexts at near-constant cost, and, under the shared recipe, retrieves beyond the trained length where our transformer variants do not; reuse carries no approximation error to bound.
For serving, retrieved passages, tool outputs, and shared documents become cacheable once, assembled per request, and removable in one memory update; open questions are whether the composite's rotary addressing holds beyond 410M, whether a fused block-local kernel closes the training-throughput gap, and how far block-local encoding carries on tasks that synthesize across many blocks.

\subsection*{AI use statement}
In this work we used generative AI tools (LLM-based coding assistants) for tasks with required disclosure: for research ideation and framing, for drafting and checking the mathematical statements and their arguments, for proposing and refining hypotheses and the pre-registered decision rules, for designing and giving feedback on experiments, for implementing the methods and evaluation harnesses, for generating the synthetic grids, for writing the corpus-preparation script, and for interpreting results.
No other required-disclosure category applies to this work.
Additionally, we used generative AI tools for tasks with recommended disclosure: writing and editing code, creating figures, drafting and editing the paper, searching and summarizing literature, formatting references, and producing simulated reviews of the manuscript.
The authors checked all AI-assisted output: code against independent re-implementations and the numerical gates (Appendices~\ref{app:gates} and~\ref{app:niah}), every reported number against its logged run, every citation against its source, and the mathematical statements themselves.
We take responsibility for the final content of this work, including text, claims, and artifacts produced with the aid of generative AI.

\subsection*{Ethics statement}
This work uses only public pretraining data (FineWeb-Edu) and involves no human subjects.
Exact removal of context-borne content is the property this architecture is designed to provide; it applies to context held in memory, not to what the weights have learned, and is not a certificate of compliance for any regulatory notion of erasure.

\bibliography{refs}
\bibliographystyle{plainnat}

\appendix

\section{Architecture and training details}
\label{app:arch}
\paragraph{Encoder.} $E$ pre-LN transformer layers with attention masked to within-block (bidirectional inside the block); a residual path from the input embeddings guarantees linear recoverability of the raw block content from $\mu_j$.
\paragraph{Reader.} $D$ decoder layers: block-local causal self-attention (within the position's own block only; attention never crosses a block boundary), cross-attention into the memory of strictly earlier blocks under a block-causal mask with a learned null row, then an FFN (hidden $4d$, GELU). Heads: $d/32$. No weight tying between embedding and head.
\paragraph{Positions and memory.} Positions: a learned table of $T$ absolute positions, indexed by (block index)$\cdot b$ + within-block offset, added to the token embeddings; encoder and reader consume the same embedded input. Memory rows are the LayerNorm of the encoder's final state, shared by all $D$ reader layers; the null row is a single learned $d$-vector prepended to the memory and projected by each layer's key/value maps; block $1$ reads the null row only.
\paragraph{Length extension.} Beyond the trained block count the slot scheme assigns indices relative to the end (\S\ref{sec:arch}); because positions enter at encode time, a block whose slot changes as the context grows is re-encoded at $O(b)$ (the $B{-}1$ most recent blocks per appended block beyond the trained length), whereas within the trained length indices are absolute and cached rows are never re-encoded. The \smem{}+RoPE composite carries no position in its rows and re-encodes nothing at any length.
\paragraph{Controls.} The Block-Attention-style arm (Appendix~\ref{app:recipe}) runs two streams over shared weights, the decoder stream taking its attention keys and values from the encoder stream's states. The encoding-tax pairs of Appendix~\ref{app:tax} differ from the deployed model: they impose a block-local \emph{causal} mask on a decoder-only model with no reader, whereas \smem{}'s encoder is bidirectional within the block.
\paragraph{Configurations and recipe.}
\begin{center}
\begin{tabular}{lccccc}
\toprule
scale & $d$ & transformer layers & \smem{} $E$ & \smem{} $D$ & peak LR \\
\midrule
160M & 768 & 12 & 3 & 7 & $6\times10^{-4}$ \\
410M & 1024 & 24 & 4 & 15 & $3\times10^{-4}$ \\
1.5B & 2048 & 24 & 4 & 15 & $2\times10^{-4}$ \\
\bottomrule
\end{tabular}
\end{center}
AdamW $\beta=(0.9,0.95)$, weight decay $0.1$ (matrices only), gradient clip $1.0$, cosine decay to $0.1\times$ peak, global batch $2^{19}$ tokens (160M/410M; at 1.5B $2^{20}$ for the transformer on 4 GPUs and $1{,}032{,}192$ for \smem{} on 3 GPUs, $28{,}611$ vs $29{,}065$ steps to the same 30B tokens), bf16 training with fp32 evaluation (TF32 matmuls; exactness gates in strict fp32), documents packed with an end-of-text separator.
Regime 1: no warmup. Regime 2: linear warmup of 2000/15641 steps at 410M and 800/6104 at 160M (${\approx}12.8\%/13.1\%$ of the horizon), identical otherwise. The depth splits follow the pilot ablation of Appendix~\ref{app:ablate}.
Validation: the last 200M tokens of the deterministic corpus stream, held out from training; each seed evaluates a fixed 0.5M-token sample of it shared across arms.
Training hardware: $7\times$H200 (single-GPU runs at 160M/410M; distributed data parallel over 4 and 3 GPUs at 1.5B).

\section{Verification gates}
\label{app:gates}
Gates G1--G8 verify the claims of \S\ref{sec:arch} on an fp64 CPU reference implementation of the architecture (deterministic kernels, freshly initialized at seed 0), independent of the production trainer; the trainer's wiring is checked by the leak test at initialization (Appendix~\ref{app:leak}) and, on the trained checkpoints, by the block-skip exactness gate (Appendix~\ref{app:serving}) and the deletion probe's forward-equivalence check (Appendix~\ref{app:niah}):
G1 analytic vs autograd cross-attention Jacobian (fp64 max deviation $<10^{-10}$);
G2 Lemma~\ref{lem:jac} bound under a memory-size sweep $N\in[16,4096]$ (no growth; large-$N$ spread flat);
G3 sharp-attention Jacobian $\to0$ monotonically;
G4 two-item worst case attains the analytic maximum at $a_1=\frac12$;
G5 exact composition: block-masked joint encoding vs per-block encoding, fp64 discrepancy $<10^{-12}$;
G6 reader invariance to memory construction form ($<10^{-12}$);
G7 bit-exact deletion: add-then-remove returns the identical tensor and identical reads;
G8 path independence: 100 random edit sequences ending at the same multiset produce bit-identical memories and outputs (after canonical row ordering).
Bit-exactness (G7/G8) holds at matched tensor shapes with canonical ordering and deterministic kernels; across kernel tilings the discrepancy is at fp64 rounding level.

\section{Theory statements and arguments}
\label{app:theory}
\paragraph{Containment.} Set $w=n$, lift the reader's window restriction, give each round its own weights (as the LM reader's untied layers do; a weight-tied recurrence simulates a universal transformer instead), initialize the $i$-th query slot to the $i$-th position's embedding, and zero the cross-attention weights: one reader round is then exactly a standard transformer layer, so $R=L$ rounds simulate an $L$-layer transformer. The reader class therefore contains the transformer as a special case with the cross-attention path unused; we do not claim strict containment, and the statement concerns this relaxation of the LM reader's layer form (full width, no window), not \smem{}'s block-local reader; the synthetic testbed's round, $Q\leftarrow Q+\beta\,\mathrm{FFN}(\mathrm{SA}(Q)+\mathrm{CA}(Q,m))$ (layer-normalized inputs, one residual branch), is a different parameterization and is not covered by it.
\paragraph{Round lower bound (sketch).} Task $\mathrm{PC}_k$: a permutation $\sigma$ on $[n]$ is stored as pairs distributed across blocks; the output is $\sigma^k(1)$. Full-width attention can pointer-double in $O(\log k)$ layers, but doubling requires all $n$ positions to hold intermediate pointers simultaneously; an adversary argument in the round-elimination style yields $R=\Omega(k)$ for narrow readers. Stated as an argument, not a fully formalized theorem. Refinement: under a one-clean-retrieval-per-slot-per-round capacity assumption, queries within a round depend only on round-start state, so on hard instances each round reveals at most one new chain point regardless of $w$---a floor $R\ge k$ flat in $w$, until $w$ is large enough to leave the assumption.
\paragraph{Matching upper construction.} Explicit weights in the same reader architecture solve $\mathrm{PC}_k$ with $R=k$ at any $w$: zero-mean orthogonal codes make LayerNorm a uniform scale; the encoder's residual path stores row $(a,\sigma(a))$; cross-attention decodes the current pointer, matches the unique source row at logit margin 50, re-encodes $\sigma(p)$; a gelu-linearized FFN turns the sum into the exact residual update. Verified $\mathrm{acc}=1.0$ exactly on all 105 cells of $w\in\{1,2,4\}\times n'\in\{8,12,16,24,32\}\times k\in\{1,2,3,4,6,8,12\}$ ($n'$ = permutation domain size), overall and on the hard bucket in the 90 cells where it is non-empty. In the trained testbed $n'{=}32$; the width-substitution threshold observed at $w{=}8$ ($w/n'{=}0.25$) marks where the capacity assumption is left (whole-table copying into slot state becomes available).

\section{Compute accounting}
\label{app:acct}
Per-run harness output:
\begin{center}
\small
\begin{tabular}{lcccc}
\toprule
scale & params (base / \smem{}) & ratio & FLOPs/token (base / \smem{}) & ratio \\
\midrule
160M & 163{,}159{,}680 / 165{,}533{,}568 & 1.0146 & $2.076\times10^8$ / $1.986\times10^8$ & 0.9565 \\
410M & 406{,}432{,}896 / 406{,}461{,}568 & 1.0001 & $7.046\times10^8$ / $6.719\times10^8$ & 0.9536 \\
1.5B & 1{,}416{,}795{,}264 / 1{,}416{,}852{,}608 & 1.0000 & $2.617\times10^9$ / $2.552\times10^9$ & 0.9750 \\
\bottomrule
\end{tabular}
\end{center}
Both ratios are \smem{}/base. FLOPs/token count every matmul at 2 FLOPs per multiply-add (attention projections, attention scores over the full context or the block, and FFN; embeddings and the LM head, identical across arms, excluded). The run logs' \texttt{acct} field uses an earlier convention that counted FFN multiply-adds once (ratios $1.0027/1.0184/1.0565$); the table uses the uniform count, under which \smem{} is $2.5$--$4.6\%$ cheaper per token because its self-attention is block-local and its reader shallower.
KV rows held per decoded context of $n$ tokens: \smem{} $D(n+b)$ (cross-attention K/V over the memory plus block-local self-attention KV) vs baseline $L\,n$; at $n{=}T{=}1024$ and our splits ($1088=n+b$) $7\cdot1088/(12\cdot1024)=0.62$ and $15\cdot1088/(24\cdot1024)=0.66$, i.e.\ $34$--$38\%$ fewer rows ($34$--$41\%$ at the 1k--8k decode contexts), driven by $D<L$ at matched parameters (the $b$-row window term is $\approx6\%$ of \smem{}'s own rows, $\approx4\%$ of the baseline's).
At full reuse the encoder pass ($15$--$22\%$ of prefill FLOPs at our depth splits and $T{=}1024$) is skipped; at partial reuse the saving accrues per cached block.
Training throughput (Regime-1 logs): at 1.5B, base $56.4$k tok/s/GPU (model FLOPs utilization $0.47$, 4 GPUs) vs \smem{} $47.7$k ($0.39$, 3 GPUs), ${\sim}15\%$ slower per GPU; on a single GPU, $23\%$ slower at 160M ($407$k vs $313$k tok/s) and $24\%$ at 410M ($169$k vs $128$k).
Iso-resource reconciliation at 1.5B (Regime 1): \smem{} reaches the transformer's final NLL at $26.4$B tokens ($88\%$ of budget), i.e.\ at $0.88\times0.975=0.86$ of the transformer's analytic FLOPs, but not within its wall-clock ($26.4/0.846=31.2$B transformer-token-equivalents $>30$B): matched analytic FLOPs favors \smem{} on this single pair; matched GPU-hours of this implementation favors the transformer.

\section{Leak tests}
\label{app:leak}
Replace every token from position 800 on by random tokens and compare the logits at positions before 799 (tolerance $10^{-4}$); the perturbation covers the encoder path (future blocks) and the reader path (future positions within block 12). Run at initialization for every run in both regimes, the change is exactly $0.0$ in all of them; the masks are structural, so the property does not depend on the parameters. A training-health canary (validation NLL must fall below the corpus unigram entropy, $7.646$ nats, by the first evaluation; deferred to the first post-warmup evaluation in Regime 2) never fired.

\section{Per-seed results and paired statistics}
\label{app:perseed}
\begin{figure}[h]
\centering
\includegraphics[width=0.5\linewidth]{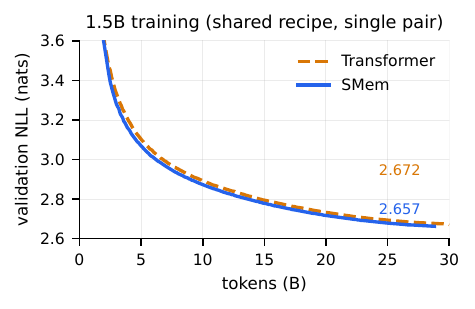}
\caption{Validation NLL over training at 1.5B, Regime 1 (single pair; the \smem{} log ends at $28.8$B tokens, its marker is the final evaluation).}
\label{fig:curve1p5b}
\end{figure}

Seed $s$ fixes data order and evaluation stream for both arms; differences are within seed pairs.
160M Regime 1 --- transformer (s0/s1/s2): 29.982 / 29.477 / 29.696; \smem{}: 28.603 / 27.980 / 28.389; paired $\Delta$: $-1.379/-1.497/-1.307$, mean $-1.394$, $t(2)=-25.2$, $p=0.002$, 95\% CI $[-1.63,-1.16]$ ppl ($-5.5\%$ to $-3.9\%$).
410M Regime 1 --- transformer: 21.23 / 21.29; \smem{}: 20.45 / 20.19; paired $\Delta$: $-0.78/-1.10$, $t(1)=-5.9$, $p=0.11$.
1.5B Regime 1 --- transformer 14.46; \smem{} 14.26 (single pair; final NLL 2.6715 vs 2.6571).
Stratified (retrievable-rare NLL, seed means): Regime 1 --- 160M $2.919$ vs $\mathbf{2.869}$, 410M $2.490$ vs $\mathbf{2.444}$, 1.5B $2.075$ vs $\mathbf{2.051}$ (\smem{} ahead at all scales); Regime 2 --- 160M $2.776$ vs $2.776$ (parity), 410M $2.382$ vs $\mathbf{2.376}$ (slightly \smem{}), against overall Regime-2 gaps of $+1.7\%/+2.1\%$ favoring the transformer.
The 1.5B training curves are in Figure~\ref{fig:curve1p5b}.

\section{The recipe control (Regime 2) and architectural controls}
\label{app:recipe}
\paragraph{Chronology.} The quality prediction of \S\ref{sec:pretraining} was registered before any pretraining run; the recipe control's decision rule was written after the Regime-1 results and before its own trigger runs---a post-hoc-motivated, prospectively-registered control.
\paragraph{Trigger (410M, single-seed arms).} Warmup-2000 at peak $3\times10^{-4}$: ppl $19.252$ ($-9.4\%$ vs the shared-recipe mean $21.26$); no warmup at $4.5\times10^{-4}$: $19.97$ ($-6.1\%$). Both exceed the registered $1\%$ re-base bar. Verification: learning-rate schedules identical across arms within each regime to $8\times10^{-17}$ absolute, gradient computation equivalent in expectation between the DDP and single-worker paths, corpus certified token-identical, all trigger checkpoints re-evaluated on a common held-out stream, and the gap present in training loss. Caveats: the warmup arm is not a single-factor variant (warmup occupies $12.8\%$ of the horizon at $98.8\%$ of the baseline's total LR budget); displayed means average per-seed eval streams (on one common stream the 410M Regime-1 baseline mean is ${\sim}21.42$; paired differences are stream-matched and unaffected); the warmup trigger run is reused as seed 0 of the re-based transformer arm.
\paragraph{Re-based comparison (tuned recipe).}
160M (3v3): transformer $26.696/26.248/26.592$ (mean $26.51$) vs \smem{} $27.118/26.712/27.040$ (mean $26.96$); paired $\Delta = +0.422/+0.464/+0.448$, mean $+0.445$, $t(2)=36.3$, $p<0.001$, 95\% CI $[+0.39,+0.50]$ ppl.
410M (2v2): transformer $19.252/18.940$ vs \smem{} $19.670/19.335$; paired $\Delta = +0.418/+0.395$, $t(1)=35.4$, $p=0.018$.
Roll gaps $0.85$--$0.97$ in every tuned run, within the Regime-1 range.
RoPE transformer: $26.11$ (Regime 1) and $25.10$ (Regime 2) at 160M; it gains $3.9\%$ from the recipe change where the learned-position transformer gains $10.8\%$. At 410M under Regime 2 (seed 0, trained on one L40S with the cluster trainer; the seed-matched local reproduction offset at 160M is $+0.2\%$) it reads $18.254$ with retrievable-rare NLL $2.362$, against the cluster's seed-0 $19.252$ (learned-position transformer) and $19.670$ (\smem{}): $-5.5\%$ and $-7.8\%$, the 160M margins ($-5.6\%$ and $-7.4\%$) carried to 410M. Under Regime 1 the same arm reads $18.886$ (retrievable-rare NLL $2.394$) against the cluster's seed-0 $21.228$ and $20.448$: $-12.4\%$ and $-8.3\%$ (160M: $-13.8\%$ and $-8.5\%$); its own recipe effect at 410M is $3.3\%$, against $9.3\%$ for the learned-position transformer and $3.8\%$ for \smem{} at seed 0.
\paragraph{\smem{}+RoPE composite (160M, 3 seeds per regime).}
Three ways of adding rotary phases to \smem{} were screened at seed 0 under Regime 2, and the winner was fixed before any further seed was trained: (v1) learned absolute positions plus rotary self-attention inside blocks, $26.69$; (v2) no absolute positions, rotary self-attention inside blocks, and in cross-attention a rotary phase applied to the query's and the memory row's \emph{block index}, so that a read sees only the block offset, $26.49$; (v3) a learned block-index embedding (slot scheme) plus rotary self-attention inside blocks, $26.65$.
v2 is the composite reported everywhere: rotary phases (base $10^4$, head dimension 32) are applied to the queries and keys of within-block self-attention by within-block offset and to the queries and keys of cross-attention by block index (the query's block, the row's block, index $0$ for the null row, which the saturating extension clamps like any other row); values are never rotated, and there is no position table. Its memory rows carry no block-position signal (within-block offsets enter only through the rotary phases of self-attention), so a cached row is reusable at any index without repositioning, and length extension is a choice made in the reader (Appendix~\ref{app:niah}).
Parameters $164.7$M, $0.5\%$ fewer than \smem{} ($165.5$M; the dropped position table) and $1.0\%$ more than the learned-position transformer ($163.2$M).
Same data order and eval stream per seed as the cluster arms, so every comparison below is seed-paired (the local trainer differs from the cluster's in one detail: weight decay is applied to the null row).
Regime 2: $26.488/26.073/26.351$ (mean $26.30$); vs the learned-position transformer $\Delta=-0.208/-0.175/-0.241$ ($-0.8\%$, $t(2)=-10.9$), vs \smem{} $-0.630/-0.639/-0.689$ ($-2.4\%$, $t(2)=-35.6$), vs RoPE $+1.229/+1.147/+1.233$ ($+4.8\%$, $t(2)=42.9$).
Regime 1: $27.175/26.765/27.080$ (mean $27.01$); vs the transformer $-2.807/-2.712/-2.616$ ($-9.1\%$), vs \smem{} $-1.428/-1.215/-1.309$ ($-4.6\%$), vs RoPE $+0.758/+0.902/+1.036$ ($+3.4\%$).
Roll gaps $0.85$--$0.89$ in all six runs.
Stratified, the composite leads every other arm on the retrievable-rare stratum under Regime 2 (per-seed NLL $2.737/2.697/2.755$ vs RoPE $2.760/2.751/2.785$, \smem{} $2.766/2.752/2.809$), so its residual deficit to the RoPE transformer lies outside the retrieval stratum, as \smem{}'s does.
Retrieval at $4\times$ length is kept when block offsets are saturated at the trained range and lost under direct rotary extension (Appendix~\ref{app:niah}); in distribution the composite reads between \smem{} and the learned-position transformer ($0.19$--$0.32$ at distance 15).
\paragraph{\smem{}+RoPE composite at 410M (Regime 2, seed 0).}
The v2 composite scaled to the 410M configuration ($d{=}1024$, 4 encoder and 15 reader layers, 32 heads of dimension 32; $405.4$M parameters, $0.25\%$ fewer than the 24-layer transformer) was trained on one L40S with the cluster trainer under the Regime-2 recipe (warmup 2000 of 15641 steps, peak $3{\times}10^{-4}$, 8.2B tokens); the run was resumed once from a checkpoint at $73\%$ after a storage fault, which re-seeds the data order from that point.
It reads $19.257$ (retrievable-rare NLL $2.367$, roll gap $0.97$): against the cluster's seed-0 learned-position transformer ($19.252$, NLL $2.384$) and \smem{} ($19.670$, NLL $2.382$) it is level with the transformer ($+0.03\%$; the seed-matched local-vs-cluster offset is $+0.05$--$0.2\%$) and $2.1\%$ below \smem{}, and it trails the RoPE transformer trained the same way ($18.254$, NLL $2.362$) by $5.5\%$, closing $29\%$ of \smem{}'s distance to it (160M: $35\%$ under Regime 2, $59\%$ under Regime 1).
As at 160M, it leads both matched arms on the retrievable-rare stratum, so the residual deficit to RoPE again lies outside retrieval; its needle readings (in distribution $0.36$ at distance 15; $0.14$ at distances 31 and 63 at $4\times$ under saturated offsets, $0.00$ under direct extension) are in Appendix~\ref{app:niah}.
Depth-matched transformer (10 layers, FFN hidden 3994, parameter ratio 1.0000): ppl $30.15/29.87/30.26$ (mean $30.09$), worse than the learned-position transformer; gap closure $-27\%$.
Block-Attention-style two-stream arm (shared-weight layers; a block-causal encoder stream providing cacheable states and a full-causal decoder stream reading it; exact parameter match): $30.13/28.94/29.90$ (mean $29.66$), at parity with the learned-position transformer, $+1.33$ ppl behind \smem{}, with the largest seed spread of any 160M arm ($1.2$ ppl); its retrieval evaluation is in Appendix~\ref{app:niah}.
\paragraph{CEPE-style arm (160M, 3 seeds, Regime 2).}
The nearest prior topology encodes chunks independently and reads them by cross-attention from a decoder whose self-attention runs over the whole prefix~\citep{cepe2024}.
We trained that reader on \smem{}'s encoder and memory: identical parameters ($165.5$M), the reader's self-attention full-causal instead of block-local, everything else unchanged, seed-paired to the Regime-2 arms.
Perplexity $29.705/29.344/29.639$ (mean $29.56$): $+9.7\%$ behind \smem{} ($+2.59/+2.63/+2.60$ ppl, every seed) and $+11.5\%$ behind the learned-position transformer; the retrievable-rare stratum is also behind ($2.898/2.869/2.920$ vs \smem{}'s $2.766/2.752/2.809$), and the roll gap falls to $0.12$ from $0.85$--$0.97$: the reader barely depends on the memory.
Two things bound the reading.
At $T{=}1024$ the full-prefix reader already sees every token the memory encodes, so the memory is redundant for this arm by construction, and the arm inherits \smem{}'s $3{:}7$ split, so its reader is a 7-layer transformer over the prefix; the comparison therefore isolates reader locality at fixed parameters and depth, not CEPE at its own operating point (chunks outside the reader's window) or its own split.
A reader-depth reference makes the second point quantitative: a 7-layer learned-position transformer with no encoder and no memory ($127.7$M parameters, same recipe, three seeds) reads $29.61/28.95/29.31$ (mean $29.29$), within $0.9\%$ of the CEPE-style arm, so under a full-prefix reader the encoded memory adds nothing that seven layers of prefix attention do not already provide, while the same seven reader layers with block-local self-attention and the memory reach $26.96$.
The arm also forfeits the block-skip path and reader-state deletion, since its reader state depends on the full prefix.
What carries quality is the block-local reader together with the memory, not the memory alone.

\paragraph{Symmetric learning-rate search (160M, local runs).}
Regime 2's learning rate ($6\times10^{-4}$) was fixed on the transformer arm, so we ran both arms at $9\times10^{-4}$, $1.2\times10^{-3}$, $1.5\times10^{-3}$, $1.8\times10^{-3}$, and $2.2\times10^{-3}$ (warmup 800, seed 0, on one L40S; the Regime-2 configuration reproduces the cluster's seed-0 runs to $+0.2\%$ for \smem{}, $27.17$ vs $27.12$, and $+0.05\%$ for the transformer, $26.71$ vs $26.70$, so the offset is small and arm-independent, the local contrast at $6\times10^{-4}$ is $+1.7\%$ as on the cluster, and local runs are compared with local runs).
Validation perplexity, \smem{} / learned-position transformer: $25.61/24.98$, $25.00/24.32$, $24.69/24.06$, $24.46/23.89$, $24.35/23.74$; a RoPE transformer at $9\times10^{-4}$ reads $23.88$; a second \smem{} seed at $9\times10^{-4}$ reads $25.24$.
The learning rate moves both arms by $10$--$11\%$ over the range, neither has turned over at $2.2\times10^{-3}$ (no run diverged; the last step buys $0.11$--$0.16$ perplexity), while the \smem{}-to-transformer contrast stays at $+2.5$, $+2.8$, $+2.6$, $+2.4$, and $+2.6\%$ (Regime 2 on the cluster: $+1.7\%$) and the RoPE lead at $7.3\%$ (Regime 2: $7.4\%$); on the retrievable-rare stratum the arms are at parity from $1.5\times10^{-3}$ upward ($2.640$ vs $2.631$ and $2.634$ vs $2.630$ nats at the two highest rates).
Warmup length matters less: \smem{} at $200/400/800/1600$ warmup steps reads $27.22/27.14/27.17/27.35$.
The recipe axis therefore moves the arms together; it does not hold an arm-specific tuning advantage for either side.

\section{Zero-shot results}
\label{app:zeroshot}
Columns: T = learned-position transformer, S = \smem{} (Regime-1 checkpoints); values are accuracies.
Regime-1 checkpoints, lm-eval-harness defaults, mean over seeds (single seed at 1.5B), per-task stderr $\le0.015$ (item-sampling only).
On ARC-e, PIQA, HellaSwag, and WinoGrande no per-scale difference exceeds about one item-sampling standard error of the difference; LAMBADA favors \smem{} by $0.027$--$0.029$ at every scale, about three standard errors (per-arm stderr $0.005$--$0.007$), and is italicized on that basis; ARC-e and HellaSwag also carry a consistent sign at smaller margins.
\begin{center}
\setlength{\tabcolsep}{4.5pt}
\small
\begin{tabular}{lcccccccccc}
\toprule
& \multicolumn{2}{c}{LAMBADA} & \multicolumn{2}{c}{ARC-e} & \multicolumn{2}{c}{PIQA} & \multicolumn{2}{c}{HellaSwag} & \multicolumn{2}{c}{WinoGrande} \\
\cmidrule(lr){2-3}\cmidrule(lr){4-5}\cmidrule(lr){6-7}\cmidrule(lr){8-9}\cmidrule(lr){10-11}
scale & T & S & T & S & T & S & T & S & T & S \\
\midrule
160M & .155 & \emph{.182} & .470 & .480 & .598 & .608 & .273 & .275 & .505 & .513 \\
410M & .231 & \emph{.260} & .523 & .537 & .637 & .631 & .293 & .297 & .502 & .509 \\
1.5B & .348 & \emph{.376} & .621 & .633 & .687 & .694 & .356 & .358 & .518 & .508 \\
\bottomrule
\end{tabular}
\end{center}

\section{Roll counterfactual (the roll gap)}
\label{app:roll}
For each evaluation batch, replace the memory each sequence reads with the memory of the neighboring sequence in the batch (a roll along the batch axis: content wrong, row statistics intact). NLL increase at final checkpoints: $0.83$--$0.85$ (160M R1), $0.94$--$0.96$ (410M R1), $1.07$ (1.5B R1); $0.85$--$0.97$ (Regime 2). A memory-ignoring model would score zero here; the gap measures how much the reader depends on its own context's memory, and it grows with scale. Addressing is tested separately by the two-key and slot-scheme probes (Appendix~\ref{app:niah}).

\section{Needle retrieval: protocol, grids, and probes}
\label{app:niah}
\paragraph{Protocol.} Haystacks are windows of FineWeb-Edu validation text. A needle = an 8-token key (mid-frequency vocabulary, ranks 200--2500) followed by a 1-token value, planted at a controlled block distance from a query (the key repeated) at the end of the context; distance is measured in blocks from the query.
$T'$ denotes the evaluation length. Metrics: exact match = argmax at the answer position equals the value; conditional gain = value log-probability under the matched key minus under a mismatched key (cancels presence bias = the value's log-probability shift from the needle's mere presence: mean $0.25$ at training length and $0.17$ at $4\times$, up to $1.0$ and $0.7$ in individual cells).
Primary harness (H200 campaign): $n{=}1024$/cell; replication harness (independent code path, validated by reproducing the logged final perplexity of every checkpoint that has one, 21 of the 33 it loads; the rest, the Regime-1 \smem{} checkpoints among them, passed the initialization leak test in their training logs): $n{=}384$/cell.
Length extension per arm in the grids: base via position interpolation, RoPE by direct extension, \smem{} via the slot scheme of \S\ref{sec:arch}; the inference-time remedies for the RoPE arm (clamp, PI, NTK, YaRN) are reported as separate rows.

\paragraph{Conditional-gain grid (primary harness, mean over seeds).}
\begin{center}
\scriptsize
\begin{tabular}{lcc cccccc}
\toprule
& & & \multicolumn{6}{c}{needle distance (blocks)} \\
\cmidrule(lr){4-9}
scale & $T'$ & arch & 1 & 3 & 7 & 15 & 31 & 63 \\
\midrule
160M & 1024 & base & 8.10 & 7.19 & 5.28 & 3.61 & -- & -- \\
160M & 1024 & \smem{} & 9.68 & 9.70 & 8.75 & 7.85 & -- & -- \\
160M & 4096 & base & 1.12 & 1.08 & 0.78 & 0.55 & 0.38 & 0.20 \\
160M & 4096 & \smem{} & 9.45 & 9.03 & 7.48 & 4.58 & 4.32 & 4.46 \\
410M & 4096 & base & 1.45 & 1.35 & 1.01 & 0.75 & 0.46 & 0.18 \\
410M & 4096 & \smem{} & 9.65 & 9.51 & 8.19 & 5.28 & 4.87 & 4.89 \\
1.5B & 1024 & base & 8.81 & 8.40 & 6.71 & 5.43 & -- & -- \\
1.5B & 1024 & \smem{} & 9.05 & 9.37 & 8.95 & 7.75 & -- & -- \\
1.5B & 4096 & base & 2.68 & 2.43 & 1.89 & 1.63 & 1.11 & 0.15 \\
1.5B & 4096 & \smem{} & 9.20 & 9.34 & 8.83 & 6.14 & 5.86 & 5.57 \\
\bottomrule
\end{tabular}
\end{center}

\paragraph{Exact-match grid (replication harness, 160M/410M, per-seed means; R1/R2 = Regime 1/2, s0 = seed 0; base = learned-position transformer, BA = Block-Attention arm, +RoPE = \smem{}+RoPE composite; sat./unsat. = saturating/unsaturated block offsets, defined below).}
\begin{center}
\scriptsize
\setlength{\tabcolsep}{2.5pt}
\begin{tabular}{ll cccc cccccc}
\toprule
& & \multicolumn{4}{c}{$T'{=}1024$, distance} & \multicolumn{6}{c}{$T'{=}4096$, distance} \\
\cmidrule(lr){3-6}\cmidrule(lr){7-12}
regime & arch & 1 & 3 & 7 & 15 & 1 & 3 & 7 & 15 & 31 & 63 \\
\midrule
R1 (160M s0) & base & .43 & .36 & .24 & .17 & .00 & .00 & .00 & .00 & .00 & .00 \\
R1 (160M s0) & \smem{} & .70 & .67 & .59 & .49 & .71 & .57 & .42 & .18 & .13 & .12 \\
R1 (160M, 3 seeds) & RoPE & .57--.61 & .37--.43 & .05--.13 & .00 & .00 & .00 & .00 & .00 & .00 & .00 \\
R1 (160M, 3 seeds) & BA & .37--.44 & .33--.36 & .25--.30 & .12--.23 & .00 & .00 & .00 & .00 & .00 & .00 \\
R2 (160M s0) & RoPE & .64 & .52 & .28 & .00 & .00 & .00 & .00 & .00 & .00 & .00 \\
R2 (160M, 3 seeds) & base & .59--.68 & .51--.57 & .36--.45 & .15--.27 & .00--.02 & .00--.01 & .00 & .00--.01 & .00--.01 & .00 \\
R2 (160M, 3 seeds) & \smem{} & .71--.80 & .65--.72 & .57--.62 & .36--.43 & .46--.74 & .24--.65 & .13--.43 & .01--.13 & .01--.11 & .01--.11 \\
R2 (410M, 2 seeds) & base & .69--.74 & .62--.68 & .51--.60 & .28--.40 & .00--.01 & .00--.01 & .00 & .00 & .00 & .00 \\
R2 (410M, 2 seeds) & \smem{} & .78--.81 & .75--.78 & .65--.68 & .46--.48 & .55--.60 & .38--.42 & .13--.20 & .01--.05 & .00--.04 & .00--.03 \\
\midrule
R2 (160M, 3 seeds) & +RoPE, sat. & .74--.75 & .66--.72 & .52--.57 & .26--.32 & .67--.73 & .47--.60 & .29--.42 & .07--.11 & .07--.12 & .05--.10 \\
R1 (160M, 3 seeds) & +RoPE, sat. & .69--.72 & .63--.67 & .49--.54 & .19--.28 & .65--.70 & .50--.55 & .24--.37 & .07--.09 & .03--.10 & .07--.08 \\
R1+R2 (160M, 6 runs) & +RoPE, unsat. & \multicolumn{4}{c}{(as above)} & .00 & .00 & .00 & .00 & .00 & .00 \\
R2 (160M s0) & +RoPE, v3 slot & .76 & .74 & .57 & .33 & .63 & .43 & .22 & .07 & .04 & .04 \\
R2 (410M s0) & +RoPE, sat. & .84 & .81 & .65 & .36 & .75 & .77 & .59 & .15 & .14 & .14 \\
R2 (410M s0) & +RoPE, unsat. & \multicolumn{4}{c}{(as above)} & .00 & .00 & .00 & .00 & .00 & .00 \\
\midrule
R1 (160M, 3 seeds) & RoPE, clamp & \multicolumn{4}{c}{(as above)} & .27--.45 & .15--.22 & .02--.05 & .00 & .00 & .00 \\
R2 (160M s0) & RoPE, clamp & \multicolumn{4}{c}{(as above)} & .39 & .22 & .05 & .00 & .00 & .00 \\
R1+R2 (160M, 4 runs) & base, clip & \multicolumn{4}{c}{(as above)} & .00 & .00 & .00 & .00 & .00 & .00 \\
R1+R2 (160M, 4 runs) & RoPE, PI & \multicolumn{4}{c}{(as above)} & .00 & .00 & .00 & .00 & .00 & .00 \\
R1+R2 (160M, 4 runs) & RoPE, NTK & \multicolumn{4}{c}{(as above)} & .00 & .00 & .00 & .00--.01 & .00 & .00 \\
R1+R2 (160M, 4 runs) & RoPE, YaRN & \multicolumn{4}{c}{(as above)} & .04--.08 & .02--.04 & .05--.09 & .04--.14 & .00--.02 & .00 \\
R1 (160M s0) & base, window & \multicolumn{4}{c}{(as above)} & .46 & .33 & .26 & .15 & .00 & .00 \\
R2 (160M, 3 seeds) & base, window & \multicolumn{4}{c}{(as above)} & .64--.69 & .47--.52 & .35--.45 & .16--.28 & .00 & .00 \\
R1 (410M s0) & base, window & \multicolumn{4}{c}{(as above)} & .68 & .54 & .46 & .29 & .00 & .00 \\
R1+R2 (160M, 4 runs) & RoPE, sink+window & \multicolumn{4}{c}{(as above)} & .56--.66 & .33--.54 & .06--.26 & .00 & .00 & .00 \\
R1+R2 (5 runs) & base, sink+window & \multicolumn{4}{c}{(as above)} & .46--.69 & .33--.53 & .26--.47 & .12--.24 & .00 & .00 \\
\bottomrule
\end{tabular}
\end{center}
``(as above)'' marks an inference-time reading of the checkpoint in the nearest model row above it; every such reading is the identity at $T'{=}T$, so those columns are unchanged. +RoPE rows are the \smem{}+RoPE composite of Appendix~\ref{app:recipe}. Its memory rows carry no position, so length extension is a choice made entirely in the reader. The composite uses \emph{saturating} block offsets (sat.): each read's offset is clipped at $B{-}1{=}15$ (key phase $\max(j,\,i{-}15)$ for query block $i$ and memory block $j$, 0-based, with the query phase $i$ unmodified), the rotary analogue of the slot scheme and the identity map at the training length. Under saturation the composite retrieves at range in all seven runs (six at 160M, one at 410M); at 410M the saturated reading gives $0.14$--$0.15$ at distances 15--63, above \smem{}'s $0.00$--$0.05$ under the same recipe, with conditional gain $4.2$--$4.6$ nats against $1.2$--$2.8$, while direct extension again reads $0.00$ everywhere.
The \emph{unsat.} rows extend the rotary offsets past the trained range instead: offsets 16--63 are never scored during training, and their scores dominate the softmax, exactly as for the RoPE transformer, so this row measures the positional scheme, not the memory. A third reading isolates which of two candidate causes is responsible: restricting each read to the last 15 blocks with unmodified phases (no untrained memory-row offset is ever scored, the null row keeping phase $0$; needles beyond 15 blocks are unreachable by construction) gives in-distribution accuracy at $4\times$ ($0.78/0.71/0.64/0.34$ and $0.82/0.72/0.58/0.33$ at distances 1/3/7/15, two seeds). Untrained-offset scores, not the number of candidate rows, account for the difference; the residual gap to the saturated reading at distances 7--15 is content competition among far blocks sharing one phase, the cost the slot scheme pays.
Its two-key probe resolves to the nearer copy in distribution ($0.41/0.19$, $0.46/0.04$, $0.28/0.00$ near/far at distances 3/7/15) and at $4\times$ up to distance 7 ($0.40/0.09$, $0.35/0.03$; at 410M $0.51/0.21$, $0.51/0.04$, and $0.08/0.11$ at distance 15). The v3 variant (learned slot embedding plus rotary inside blocks) retrieves at range under its native slot scheme without any eval-time choice.
The last three rows give the transformer arms a saturating extension of their own: \emph{RoPE, clamp} clips every relative offset at $T{-}1$ (the token-level analogue of saturated block offsets; the identity at $T$ up to $2\times10^{-5}$ in strict fp32), and \emph{base, clip} keeps native positions for the last $T$ tokens and assigns position $0$ to everything older, the token-level slot scheme. Clamping recovers part of the near range for RoPE and nothing at $\ge15$ blocks; clipping leaves the learned-position arm at $0.00$ everywhere, since $3072$ tokens then share one position. Inference-time length-extrapolation remedies for the RoPE arm, applied without fine-tuning and each the identity at $T$ (verified in strict fp32): position interpolation (\emph{PI}, positions scaled by $T/T'$) reads $0.00$ at every distance; NTK-aware base rescaling (\emph{NTK}, base $\times\,4^{\,d_h/(d_h-2)}$) reads $0.00$--$0.01$; YaRN (\emph{YaRN}, per-dimension ramp between interpolation and extrapolation with $\beta_{\text{fast}}{=}32$, $\beta_{\text{slow}}{=}1$ and its logit temperature $0.1\ln 4+1$) recovers part of the near range ($0.02$--$0.14$ within 15 blocks, conditional gain $1.4$--$5.1$ nats) and nothing beyond it ($0.00$--$0.02$ at 31 blocks, $0.00$ at 63; conditional gain $\le1.2$ and $\le0.12$ nats), the same shape as the clamp.
The remaining transformer option, \emph{base, window}, reads only the last $T$ tokens with native positions: it reproduces in-distribution accuracy at $4\times$ up to distance 15 (the needle is then $1012$--$1020$ tokens back, just inside the window) and reads $0.00$ at distances 31 and 63 (conditional gain within $\pm0.3$ nats of zero, five runs), which the window cannot reach. A StreamingLLM-style eviction reading (\emph{sink+window}: 4 sink tokens plus the last $T{-}4$ tokens, the retained tokens re-positioned contiguously as in a streaming cache) behaves like the window: RoPE reads $0.56$--$0.66/0.33$--$0.54/0.06$--$0.26$ at distances 1/3/7 and $0.00$ from 15 blocks on, the learned-position arm $0.46$--$0.69$ down to $0.12$--$0.24$ at 15 and $0.00$ beyond; an eviction policy preserves what its window holds and nothing farther. This is why the range claims of \S\ref{sec:niah} are stated at $d{\ge}31$: at $d{\le}15$ a windowed transformer ties or beats \smem{} at $4\times$ under either recipe, and the separation there is one of re-indexing, not reach.
\paragraph{A fully pretrained RoPE model on the same probe.} Pythia-410M (RoPE, context 2048, ${\sim}300$B tokens), same protocol with text re-tokenized to its vocabulary and an 8-token key of mid-vocabulary tokens, $n{=}384$ per cell. In distribution it retrieves: exact match $0.78/0.73/0.52/0.64$ at distances $1/3/7/15$ ($T'{=}1024$) and $0.84/0.71/0.58/0.47/0.77$ at $1/3/7/15/31$ ($T'{=}2048$). At $T'{=}4096$, twice its context, under direct rotary extension it reads $0.00$ at every distance ($1$--$63$; conditional gain within $\pm0.2$ nats of zero). So the in-distribution RoPE zero of our 3.2B-token arms reflects their training budget, whereas collapse beyond the trained context under direct extension is shared by a model trained two orders of magnitude longer.
The primary harness's cross-scale summary at $T'{=}4096$, $d\in\{31,63\}$ (beyond any $T$-token window) is $0.14$--$0.28$ (\smem{}: $0.14/0.15$, $0.25/0.22$, $0.28/0.28$ at 160M/410M/1.5B) vs $0.00$--$0.02$ (base); at $d{=}15$ it reads $0.13/0.25/0.26$ vs $0.00$--$0.02$; the replication harness reads the 160M cells at $0.13/0.12$ ($d{=}31/63$) and $0.18$ ($d{=}15$).
Under Regime 2 the far-range margin survives in nats where exact match sits near the floor: conditional gain at $d{=}31/63$ is $1.2$--$3.2$ nats for \smem{} ($2.93/3.21$, $2.15/2.34$, $1.56/1.19$ at 160M; $2.61/2.57$, $1.23/1.31$ at 410M) against $0.00$--$0.49$ for the learned-position transformer.
BA conditional gain at $T'{=}4096$ is $\le0.35$ nats at every distance (three seeds).
\paragraph{Deleting the needle's block on the pretrained models.} Theorem~\ref{thm:edit} says the post-deletion memory is the never-encoded one; this probe measures the behavior that follows. On \smem{} at 160M, 410M, and 1.5B (Regime 1; Regime 2 at 160M/410M), $T'\in\{1024,4096\}$, distances $1$--$63$, $n{=}384$ per cell, we plant the needle and then remove the needle block's rows from memory, with the remaining blocks keeping their positions (the final block's own rows, never read at the query, are dropped in every condition; that reading equals the standard forward to $\le1.1\times10^{-5}$ in fp32). Exact match goes from $0.03$--$0.82$ (intact, by cell) to $\le0.003$ in all 40 cells (at most one hit in 384; the never-planted rate is $0.00$ in every cell); the value's log-probability after deletion sits within $0.0$--$0.7$ nats of the never-planted reference (the same context queried with a key that was not planted; $-11.0$ to $-12.0$ nats; mean over cells $-0.3$, i.e.\ slightly below it, consistent with the encoder still having seen the key in the query block); and deleting a neighboring, non-needle block instead leaves accuracy at the intact value (within $0.04$). Deletion removes the retrieval and only the retrieval.

\paragraph{Two-key probe.} Two identical keys, the nearer at distance 1 and the farther at distance 15, query at the end ($T'{=}T$, $n{=}1024$): \smem{} argmax returns the near value / far value at $0.65/0.01$ (160M), $0.66/0.03$ (410M), $0.50/0.11$ (1.5B). The far rate sits below the single-key rate at distance 15, so the nearer copy is actively preferred rather than the far one merely being harder; the probe measures preference by distance and does not separate recency from slot position.

\paragraph{Natural-text variant.} Mining validation cases where a rare token's earlier occurrence sits in a strictly earlier block and ablating that source: at 1.5B, $T'{=}4096$, \smem{} $\Delta$acc $+0.36/+0.23/+0.17/+0.12$ at distance bins $1/2\!-\!3/4\!-\!7/8\!-\!15$ ($n{=}215/111/70/49$) vs base $+0.22/+0.23/+0.09/+0.06$; a transformer reading only the last $1024$ tokens ($+0.41/+0.30/+0.19/+0.12$) ties or beats \smem{} in every bin inside its window and is blind beyond it.

\paragraph{Open-book QA.} TriviaQA closed/open + SQuAD-v1, $n{=}1500$, greedy 24-token decode, two harnesses. Harness A point estimates favored \smem{} (1.5B TriviaQA open-minus-closed delta $+0.155$ vs $+0.103$); harness B (per-item bootstrap) gives difference-in-differences $+0.003/+0.006/-0.006$ at 410M-R1/1.5B-R1/410M-R2 with every 95\% interval including zero, while SQuAD-v1 exact match favors \smem{} by $+0.012/+0.011/+0.031$ (two of the three intervals exclude zero) where harness A's F1 is mixed in sign. The QA contrast is harness-sensitive and not resolved under item resampling at these scales. The capability evidence of this section rests on the needle, distance, and recency probes.

\section{Serving measurements}
\label{app:serving}
All serving benchmarks: bf16, both arms in the same implementation; block-skip, delete-then-generate and deletion-latency cells on a single H200, decode cells on a single L40S (48 GB, $864$ GB/s nominal); the transformer prefill/suffix-recompute arm uses the flash lower-right-causal kernel (fastest available).
\paragraph{Block-skip exactness and TTFT.} Final-block logits, block-skip vs full reader pass: max abs $4.7\times10^{-5}$, max relative $2.4\times10^{-6}$, argmax agreement $100\%$, all 12 scale$\times$length cells (four lengths, $1$k--$8$k; strict fp32). Block-skip cost $3.1$--$6.2$\,ms across lengths (batch 1) vs cold prefill growing with context (Figure~\ref{fig:cache}b; the 410M cold-prefill cell reads the same at 8k as at 4k, an anomaly we did not resolve); the percentage saving ($67$--$69\%$ at training length, $88$--$94\%$ at $4$--$8\times$) grows mechanically with length, so absolute costs are the informative quantity.
\paragraph{Decode.} KV-cached incremental decode with caches preallocated and written in place for both arms (no per-step cache growth), 128 decoded tokens after 66 warmup steps, median of 3 repetitions; exactness-gated against the native forward in strict fp32 across a block boundary at batch 1 and 8 (max err $2.2\times10^{-4}$, argmax disagreement 0). Grid: 3 scales $\times$ 3 arm/K-V configurations (transformer; \smem{} with materialized cross-attention K/V; \smem{} recomputing K/V from the memory rows) $\times$ contexts $\{1$k$,4$k$,8$k$\}$ $\times$ batch $\{1,16,64\}$, cells whose estimated footprint exceeded a $34$ GB budget skipped. Decode contexts beyond $T$ assign slots by clamping at $B$ (blocks after the $B$-th share the last slot), which appends without re-encoding; this changes position ids only, not the work per token. A cell is \emph{launch-bound} when its per-token time is within $1.5\times$ of the same arm's batch-1 time; batch-1 cells and batched cells below $590$ GB/s measure kernel dispatch rather than the architecture and are reported without interpretation. The cells below have both arms at ${\ge}590$ GB/s of achieved traffic (K/V plus the bf16 weight copy) and are treated as bandwidth-bound; by the per-token rule three are borderline (160M/8192/16 and 410M/1024/64 on the \smem{} side, 410M/4096/16 on both), and dropping them leaves the ratio at $1.44$--$1.67\times$:
\begin{center}
\scriptsize
\begin{tabular}{llcccccc}
\toprule
scale & ctx / batch & transf. tok/s & \smem{} tok/s & ratio (\smem{}/transf.) & transf. GiB & \smem{} GiB & \smem{} recompute tok/s \\
\midrule
160M & 4096 / 64 & 4261 & 6969 & $1.64\times$ & 10.8 & 10.4 & 980 \\
160M & 8192 / 16 & 2006 & 3241 & $1.62\times$ & 6.3 & 5.8 & 634 \\
160M & 8192 / 64 & 2278 & 3810 & $1.67\times$ & 20.3 & 19.5 & --- \\
410M & 1024 / 64 & 5135 & 7347 & $1.43\times$ & 9.8 & 8.2 & 1060 \\
410M & 4096 / 16 & 1397 & 2018 & $1.44\times$ & 9.1 & 7.6 & 292 \\
410M & 8192 / 16 & 796 & 1212 & $1.52\times$ & 15.7 & 12.8 & 199 \\
1.5B & 1024 / 64 & 2528 & 3636 & $1.44\times$ & 23.0 & 19.8 & --- \\
1.5B & 4096 / 16 & 675 & 1010 & $1.49\times$ & 21.6 & 18.6 & 88 \\
\bottomrule
\end{tabular}
\end{center}
The throughput ratio tracks the KV-row ratio ($0.62$--$0.66$ rows, hence $1.5$--$1.6\times$ expected); peak memory includes fp32 weights and their bf16 copies, so the $34$--$38\%$ KV saving appears as $4$--$18\%$ of the total. Launch-bound cells: batch 1 gives $1.03$--$1.21\times$ at all scales and contexts. The recompute regime avoids the materialized K/V ($2.1$ vs $5.8$ GiB at 160M/8192/16) but re-projects every memory row per layer per step and is compute-bound, at $0.09$--$0.20\times$ the materialized throughput in the batched cells above.
\paragraph{Delete-then-generate.} \smem{} (row-slice + block-skip read) vs transformer (suffix KV recompute + next token); the deleted block sits at $25/50/75\%$ of the context, and the transformer's cost grows with the suffix after it, so the advantage is smallest when deleting near the end and largest near the start. Grid: 512 and 4096 blocks, all scales (18 cells): $8.5\times$ (160M, 512 blocks, deletion nearest the end) to $452\times$ (160M, 4096 blocks, deletion nearest the start); 1.5B at 4096 blocks: $125$--$126$\,ms vs $13.9$--$29.9$\,s ($111$--$237\times$).
\paragraph{Decode under the index-mapped layout.} The decode table above stores memory contiguously; the index-mapped layout that keeps deletion flat (below) reads every row through the map. With a naive per-step gather (an \texttt{index\_select} of the cross-attention K/V per layer, no fused kernel) the exactness gate is unchanged ($\le5.7\times10^{-5}$, argmax agreement $100\%$) and per-token time in four of the five batch-16 cells rises to $1.8$--$2.8\times$ the contiguous layout's ($7.33$ vs $3.67$\,ms at 160M/4096/16; $13.95$ vs $4.94$\,ms at 160M/8192/16; $20.7$ vs $7.9$\,ms at 410M/4096/16; $13.8$ vs $7.7$\,ms at 1.5B/1024/16; the fifth, 410M/1024/16, is launch-bound and reads $1.07\times$), which is $0.55$--$0.74\times$ the transformer's decode speed rather than $1.4$--$1.7\times$. The gather is the indirection that paged-attention kernels perform without a copy; our implementation has no such kernel, so the flat-deletion layout and the decode advantage are, in this implementation, measured under different layouts.
\paragraph{Deletion latency sweep.} Memory-update cost to $65{,}536$ blocks, contiguous vs index-mapped layouts (Figure~\ref{fig:cache}a): contiguous grows from $512$ blocks ($0.03$\,ms there, $0.21$\,ms at 4k, $3.0$\,ms at $65{,}536$ blocks); index-mapped stays at $0.014$--$0.027$\,ms but pays a per-read gather growing with memory ($3.3$\,ms at $65{,}536$ blocks vs $0.003$\,ms contiguous)---the layouts trade delete-time against read-time cost.

\section{Memory-suppression decomposition and the deletion task}
\label{app:suppress}
These experiments use the pilot-phase masked-recovery architecture (bidirectional reading, so own-block memory reads exist; the causal LM of \S\ref{sec:arch} reads strictly earlier blocks only), $d{=}256$, 3 seeds, trained on a synthetic curriculum whose sequences contain rare tokens copied across blocks; the deletion test set holds one target per case with all evidence blocks identified.
Copy-task accuracy under progressively suppressed memory access: intact $0.597$; own-block memory only $0.365$; no cross-attention at all $0.189$ (which also removes own-block reads; the locally-predictable control stratum reads $0.426/0.419/0.366$ under the same three conditions---it uses memory weakly, and is unchanged only under \emph{deletion}, the relevant operation).
The cross-block share ($0.597-0.365=0.23$) is what deleting the evidence blocks removes in \S\ref{sec:delete}: with every evidence block deleted, target accuracy falls from $0.58$--$0.61$ (three seeds) to $0.361$--$0.363$, at the own-block-only level, while the control stratum changes by at most $0.001$ ($0.428\to0.429$ in one seed).

\section{Encoding-tax measurements}
\label{app:tax}
\paragraph{From-scratch anchor (primary).} Parameter-identical ${\approx}83.6$M models ($d{=}512$, 10 layers; only the attention mask differs: block-local-causal vs full-causal), 1B FineWeb-Edu tokens, three seed pairs. Per-pair (tax / retrievable-rare / non-retrievable-rare, nats): $0.195/{+}1.403/{-}0.088$; $0.132/{+}1.317/{-}0.170$; $0.140/{+}1.291/{-}0.155$. Mean $\pm$ SD: tax $0.156\pm0.034$, retrievable $+1.337\pm0.059$, non-retrievable $-0.138\pm0.044$ (same sign in all pairs). Concentration (per-pair retrievable/overall): $7.2/10.0/9.2\times$, mean $8.8\times$.
\paragraph{Pretrained-model upper bound.} Pythia-410M, WikiText-2, a 64-token block-causal mask imposed at inference ($T{=}512$; the model was never trained with it), both conditions under identical explicit causal masks; full-attention sanity NLL 3.29. Tax $+3.04$ nats on tokens whose target occurs in an earlier block, $+0.57$ elsewhere.
\paragraph{Controlled decomposition (pilot scale).} A bare block-local model (no reader) pays a $15$--$21$ point accuracy deficit on cross-block-dependent predictions across encoder widths $d\in\{64,128,256\}$ ($+15.1/+19.9/+20.8$), which does not shrink with width; with a reader of four rounds on both arms the deficit is gone ($\Delta\le0$ at every width), while an equal-FLOPs deepening of the bare encoder (a fifth layer at $d{=}256$, returning the full-attention arm's extra FLOPs as depth) moves accuracy by $+0.002$. The comparator throughout is the unmasked-encoder counterpart, not the transformer of \S\ref{sec:pretraining}.

\section{Synthetic resource-relation grids}
\label{app:synth}
Task $\mathrm{PC}_k$ (permutation domain $n'{=}32$), reverse-retrieval-aware hard subset (effective hops $=k$), vocabulary-ladder curriculum, logit masking to active ids; testbed reader = weight-tied rounds (Appendix~\ref{app:theory}).
Seed accounting for the $w\le4$ regression (30 points over $k\in\{1,2,3,4\}$, the $k{=}1$ control row included): $k{=}1$: 5 seeds by design (control row, all at floor); $k{=}2$: 10 seeds---5 reached the floor at $w\le4$ (one of these also succeeded at $w{=}8$ with $R{=}1$, two more at $w{=}16$), 3 succeeded only at $w{=}8$ with $R{=}1$ (width substitution), and 2 did not succeed within budget ($5+3+2=10$); $k{=}3,4$: 10 each, all formed.
Per-seed minimal $R\cdot w$ at $w\le4$: $k{=}1$: all 1; $k{=}2$: all 2; $k{=}3$: $[3,3,3,4,4,4,4,6,6,16]$; $k{=}4$: $[4,4,4,4,4,4,8,8,8,8]$.
Per-width floor: at $w\le4$ the search ran $R\ge k$ only, so the floor is not tested from below there; among formed cells the minimal successful $R$ equals $k$ exactly, and sub-$k$ rounds were run only at $w\ge8$ ($k{=}2$, $R{=}1$), where they succeed by width substitution.
Robustness: OLS slope $1.270$, 95\% CI $[0.98,1.56]$; Theil--Sen $1.00$; leave-one-out range $[1.24,1.29]$; without the $k{=}3$ outlier: $1.24$ $[1.03,1.44]$; minima over all $w$: $1.15$ $[0.77,1.54]$; worst-case censoring (five non-qualifying $k{=}2$ seeds imputed at the sweep maximum $R\cdot w{=}8$): $1.09$ $[0.67,1.51]$---under this imputation the logarithmic alternative is not excluded, so that exclusion is conditional on the achieved-minima analysis.
$k{=}6$: 88 cells over a registered ${\sim}19$ GPU-hour budget produced zero formations; budget-censored and uninformative about the floor (with zero successes, the absence of sub-$k$ successes is vacuous).

\section{Pilot-scale split ablation}
\label{app:ablate}
Pilot scale ($d{=}256$): adding reader layers at a fixed 2-layer encoder ($D\in\{2,4,6\}$, $8.0\to12.2$M parameters) improves perplexity monotonically ($6.40\to5.79\to5.52$), deepening the encoder at fixed reader depth changes perplexity within noise, and smaller blocks help at fixed compute ($5.74$ at $b{=}16$ vs $5.86$ at $b{=}64$). The production split ($E{:}D = 3{:}7$ / $4{:}15$) follows this ablation and is held fixed across scales.

\section{Training practices and precedents}
\label{app:pathologies}
Items (1)--(3) sharpen effects with published precedents~\citep{eureka2024}; the rest are standard practice. Symptom $\to$ diagnosis $\to$ fix, from the pilot phase:
(1) long warmup at synthetic formation time prevents retrieval circuits from forming (threshold near 300 steps; the Regime-2 runs show pretraining-scale \smem{} tolerates standard warmup, so this binds the synthetic setting);
(2) over-large curriculum vocabulary steps collapse multi-hop formation---small ladder steps with per-stage optimizer reset;
(3) weight-tied recurrence at $R>k$ starves early-round parameters under naive truncated backprop---random-depth truncation with a straight-through initial state;
(4) embedding-scale rules are optimizer-specific---standard $0.02$ init with AdamW;
(5) inactive-vocabulary logits drift in curricula---mask logits to active ids;
(6) positional signal far below content scale leaves the addressing channel unused---balance embedding norms;
(7) LayerNorm first inside a residual branch attenuates gradients---zero-init the branch output;
(8) aggregate metrics hide stratum effects---decompose by retrievability stratum and distance;
(9) fp16 overflows on small models---bf16 with fp32 evaluation.

\end{document}